\documentclass[journal,dvipsnames,svgnames,table]{IEEEtran}

\usepackage[sort&compress,numbers,sectionbib]{natbib}
\usepackage{multicol}
\usepackage[bookmarks=true]{hyperref}
\usepackage{amsmath,amsfonts,mathtools,amsthm}
\usepackage{upgreek}
\usepackage{tikz}
\usepackage{braids}
\usepackage{subcaption}

\renewcommand{\dotsb}{%
  \mathinner{\cdotp\mkern-3mu\cdotp\mkern-3mu\cdotp}%
}

\newcommand{\mydots}{\scalebox{0.3}{$\dotsb$}}
\usepackage{amssymb}  
\usepackage{pifont}
\usepackage{cite}
\usepackage{epstopdf}
\usepackage{float}
\newfloat{algorithm}{t}{lop}
\usepackage{stfloats}
\usepackage{mathtools}
\usepackage{algorithm}
\usepackage{algorithmicx}
\usepackage[noend]{algpseudocode}

\let\labelindent\relax
\usepackage{enumitem}

\newtheorem{remark}{Remark}
\theoremstyle{remark}

\usepackage{calrsfs}
\DeclareMathAlphabet{\pazocal}{OMS}{zplm}{m}{n}
\DeclareMathAlphabet\mathbfcal{OMS}{zplm}{b}{n}

\usepackage{balance}

\newcommand{\figref}[1]{Fig.~\ref{#1}}

\usepackage{bm}

\usepackage{url}

\newcommand{\xxnote}[3]{}
\ifx\hidenotes\undefined
  \usepackage{color}
  \renewcommand{\xxnote}[3]{\color{#2}{#1: #3}}
\fi

\begin{document}

\title{Implicit Multiagent Coordination\\
at Unsignalized Intersections via Multimodal Inference Enabled by Topological Braids}

\author{Christoforos Mavrogiannis,~\IEEEmembership{Member,~IEEE,}
        Jonathan A. DeCastro,~\IEEEmembership{Member,~IEEE,}
        and~Siddhartha~S.~Srinivasa,~\IEEEmembership{Fellow,~IEEE}
\thanks{This work was partially funded by the National Institute of Health R01 (\#R01EB019335), National Science Foundation CPS (\#1544797), National Science Foundation NRI (\#1637748), the Office of Naval Research, the RCTA, Amazon, and Honda Research Institute USA.}
\thanks{C. Mavrogiannis and S. S. Srinivasa are with the Paul G. Allen School of Computer Science \& Engineering, University of Washington, Seattle, WA 18195, USA (e-mail: cmavro@cs.washington.edu, siddh@cs.washington.edu).}
\thanks{J. A. DeCastro is with the Toyota Research Institute, Cambridge, MA 02139, USA (e-mail: jonathan.decastro@tri.global).}
}

\maketitle

\begin{abstract}
We focus on navigation among rational, non-communicating agents at unsignalized street intersections. Following collision-free motion under such settings demands nuanced \emph{implicit} coordination among agents. Often, the structure of these domains constrains multiagent trajectories to belong to a finite set of \emph{modes}. Our key insight is that empowering agents with a model of these modes can enable effective coordination, realized implicitly via intent signals encoded in agents' actions. In this paper, we represent modes of joint behavior in a compact and interpretable fashion using the formalism of topological \emph{braids}. We design a decentralized planning algorithm that generates actions aimed at reducing the uncertainty over the mode of the emerging multiagent behavior. This mechanism enables agents that individually run our algorithm to collectively reject unsafe intersection crossings. We validate our approach in a simulated case study featuring challenging multiagent scenarios at a four-way unsignalized intersection. Our model is shown to reduce frequency of collisions by $>$65\% over a set of baselines explicitly reasoning over trajectories, while maintaining comparable time efficiency.
\end{abstract}


\IEEEpeerreviewmaketitle

\begin{figure}[t]
\centering
\begin{subfigure}{0.48\linewidth}
\centering
\includegraphics[trim = {0cm 0cm 0cm 0cm}, clip, width = \linewidth]{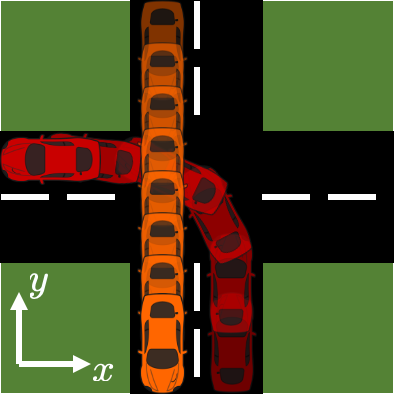}
\caption{The red agent accelerates, passing before the orange agent.\label{fig:fig3}}
\end{subfigure}
~
\begin{subfigure}{0.48\linewidth}
\centering
\includegraphics[trim = {0cm 0cm 0cm 0cm}, clip, width = \linewidth]{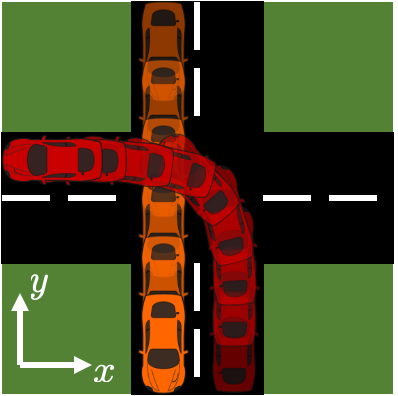}
\caption{The orange agent accelerates, passing before the red agent.\label{fig:fig4}}
\end{subfigure}
\\
\begin{subfigure}{0.48\linewidth}
\centering
\includegraphics[trim = {0cm 0cm 0cm 0cm}, clip, width = \linewidth]{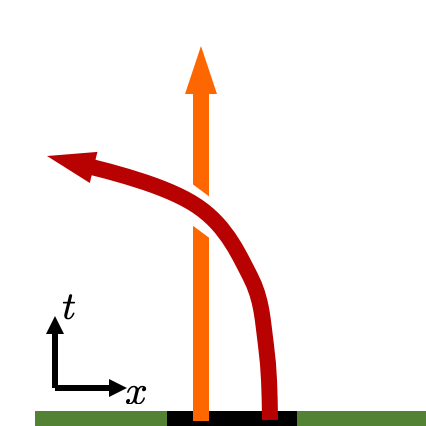}
\caption{Trajectories of agents in \figref{fig:fig3}, plotted in spacetime.\label{fig:leftbraid}}
\end{subfigure}
~
\begin{subfigure}{0.48\linewidth}
\centering
\includegraphics[trim = {0cm 0cm 0cm 0cm}, clip, width = \linewidth]{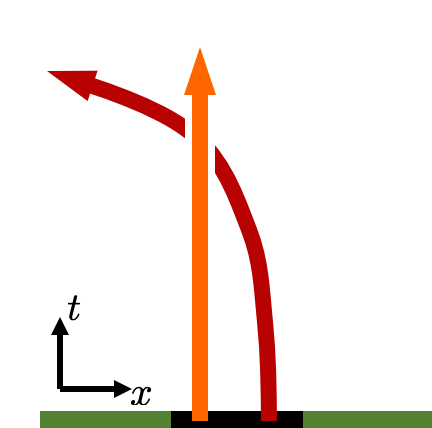}
\caption{Trajectories of agents in \figref{fig:fig4}, plotted in spacetime.\label{fig:rightbraid}}
\end{subfigure}
\caption{Two agents cross an unsignalized intersection. The red agent starts from the bottom and moves to the left whereas the orange agent starts from the top and moves to the bottom. Figs (\subref{fig:fig3}) and (\subref{fig:fig4}) depict two different executions representing distinct modes of crossing the intersection. Spacetime projections of the two executions, as depicted in Figs (\subref{fig:leftbraid}) and (\subref{fig:rightbraid}) result in different trajectory crossings revealing topologically distinct properties. In this paper, we identify topologically distinct executions using the theory of topological braids \citep{artin}.\label{fig:motivation}}
\end{figure}

\section{Introduction}\label{sec:introduction}

Street environments such as intersections often feature a significant amount of spatial structure, such as crosswalks, sidewalks or dedicated lanes. However, due to driver-to-driver variability, local customs and inconsistencies in the placement of signs and traffic lights, they do not always feature concrete mechanisms for organizing traffic flows temporally.  For instance, street intersections lacking traffic lights and signage (see \figref{fig:motivation}) is a situation (resulting from e.g. accidents, construction, loss of power, inadequate signage) most drivers have encountered, and is, in fact, prevalent in developing countries \citep{patil_microscopic_2016}.
Uncertainty could be reduced through the use of standard means of signaling, such as turn signals, horns, or even gaze, gestures and verbal communication.  However, since such signaling is inconsistent, relying on these cues for safety comes at a significant risk to safety.  On the other hand, in terms of risk of collision, uncertainty can be reduced by considering the trajectory of the other vehicles.  Consider, for instance, a hurried or inattentive driver approaching a busy intersection.  We posit that reacting to cues that such a driver has not yet slowed down upon approaching the intersection gives valuable insight into efficiently coordinating the remaining agents' future actions to ensure that they are able to safely traverse the intersection without rendering their behaviors to be overly conservative and less efficient in other situations.

The effects of wrongly responding at intersections may vary from inefficient intersection crossings (traffic backups) to catastrophic situations involving collisions. For reference, in the United States, during the year 2018, 43.7\% of all motor vehicle crashes occurred at intersections (2,943,717 out of 6,734,416 incidents). Out of these, 8,245 incidents involved fatalities, representing the 24.5\% of all fatal crashes for the same year (out of a total of 33,654 fatal crashes) \citep{fatality}. While the circumstances of each accident may differ, we view the high uncertainty and lack of timely coordination as major contributing factors to this sad reality.




\begin{figure*}
\centering
\begin{subfigure}{0.23\linewidth}
\centering
\includegraphics[trim = {.5cm 0cm 0.2cm 0.1cm}, clip, width = \linewidth]{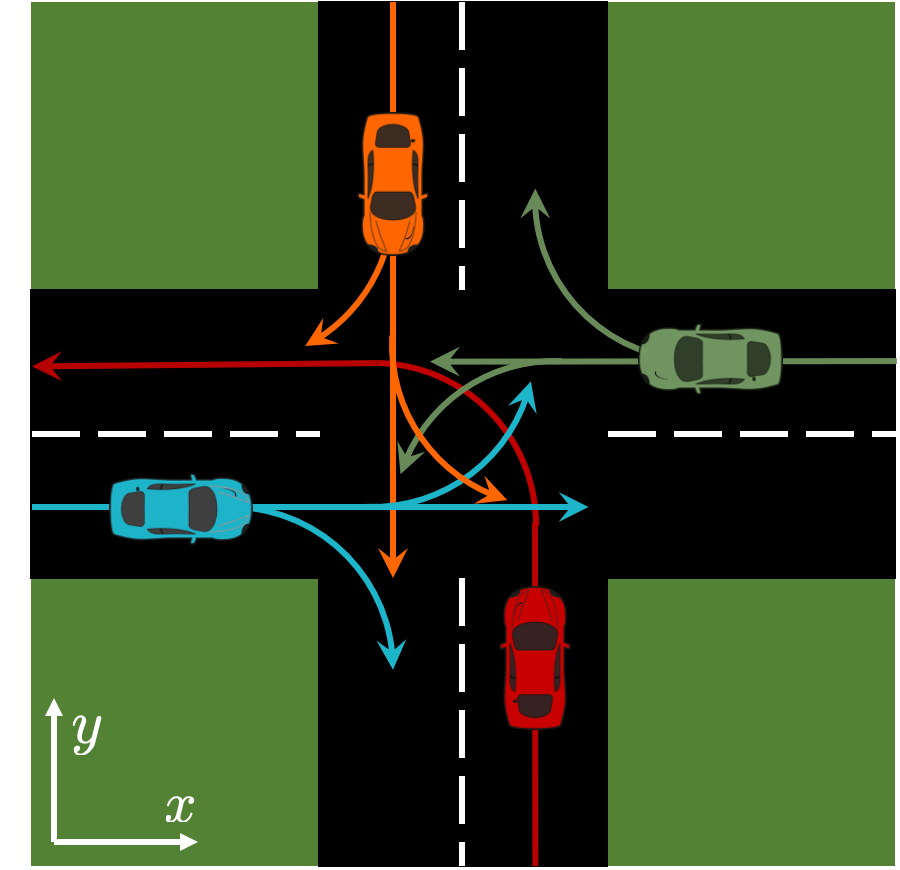}
\caption{The red agent is uncertain of the intended paths of other agents. This uncertainty complicates its decision making.\label{fig:allpaths}}
\end{subfigure}
~
\begin{subfigure}{0.23\linewidth}
\centering
\includegraphics[trim = {0cm 0cm 0.2cm 0.1cm}, clip, width = \linewidth]{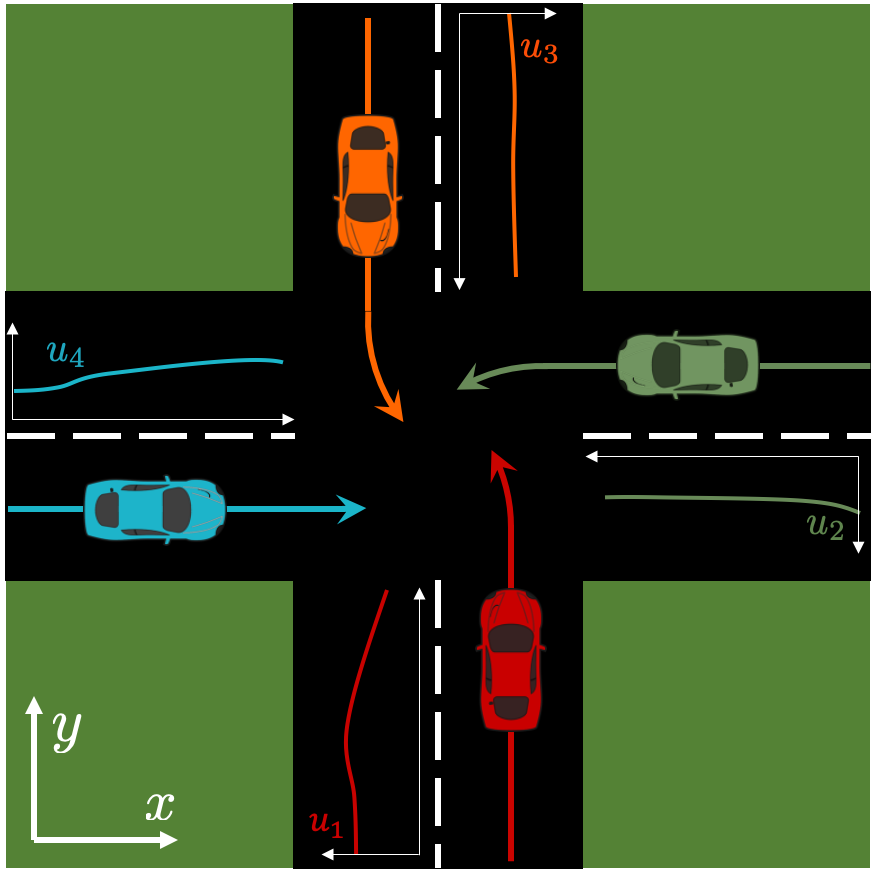}
\caption{Agents' control profiles so far (depicted as graphs with time) are indicative of their intentions. \label{fig:controldiagrams}}
\end{subfigure}
~
\begin{subfigure}{0.23\linewidth}
\centering
\includegraphics[trim = {.9cm 0cm 0.6cm 0cm}, clip, width = \linewidth]{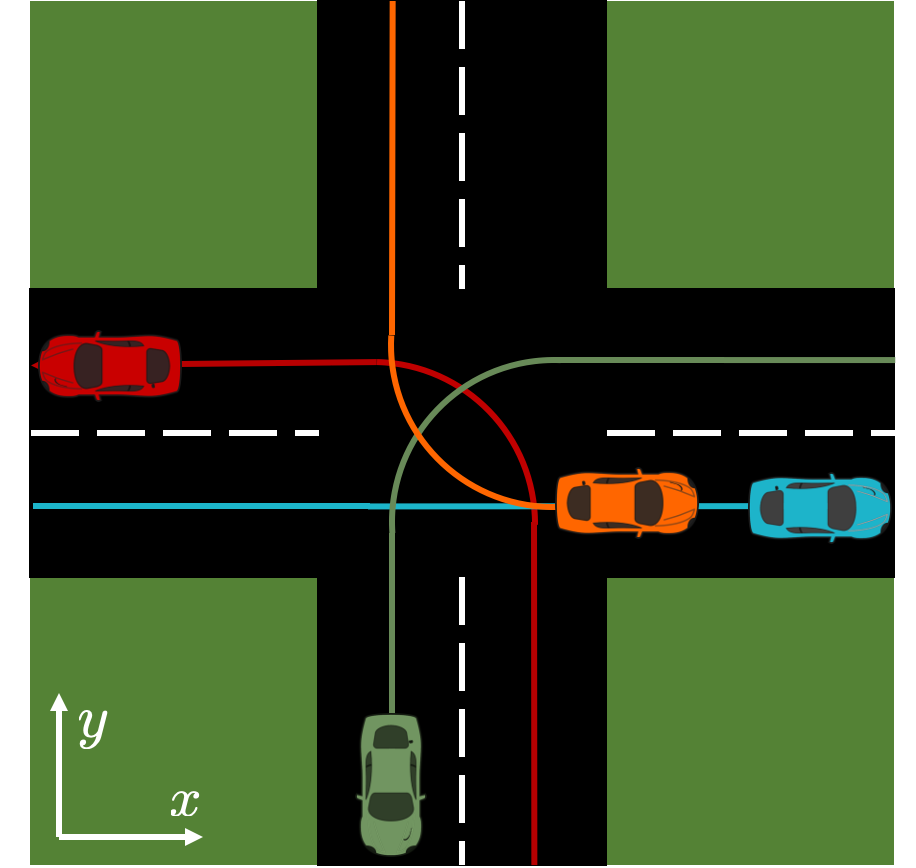}
\caption{Top view of the trajectories that agents followed from the beginning to the end of the intersection. \label{fig:pathsfollowed}}
\end{subfigure}
~
\begin{subfigure}{0.23\linewidth}
\centering
\includegraphics[width = \linewidth]{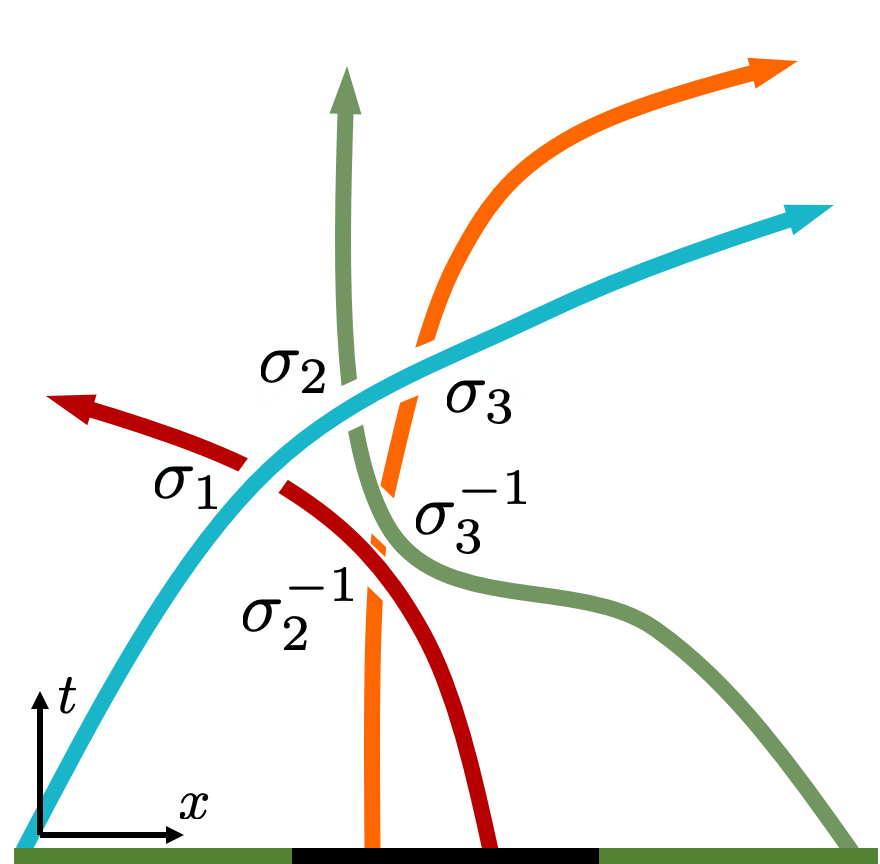}
\caption{Projection of agents' trajectories onto the $x$-$t$ plane and topological characterization of emerging trajectory crossings. \label{fig:spatiotemporal}}
\end{subfigure}
\caption{The ego agent (red car) enters an unsignalized intersection. Despite unaware of others' destinations (\subref{fig:allpaths}), by observing the joint state history (\subref{fig:controldiagrams}), the ego agent may infer the intentions of others. Based on the inferred intentions, the ego agent may further infer the upcoming sequence of intersection crossings (\subref{fig:pathsfollowed}). The trajectory crossings unveiled in the 2-dimensional projection of \figref{fig:spatiotemporal} reveal crucial properties of the collision-avoidance protocol followed by agents. The formalism of topological braids \citep{artin} allows us to describe the execution of the complete execution of (\subref{fig:pathsfollowed}) as a word composed of symbols (Fig. (\subref{fig:spatiotemporal})) representing pairwise topological relationships among agents' trajectories.\label{fig:introfigs}}
\end{figure*}

With the projected advent of autonomous vehicles \citep{McKinsey}, there is a big opportunity for reducing the number of catastrophic failures described above. In particular, we find that the complications arising from the lack of explicit coordination could be reduced by developing and leveraging mechanisms of \emph{implicit} communication. As a first step to making our case, we focus on the model setup of an unsignalized four-way street intersection where multiple rational\footnote{Agents that intend to follow short and collision-free paths.}, non-communicating but perfectly-observing agents navigate in close proximity. This setup features the key characteristics identified above: a) the domain features no concrete traffic rules to regulate traffic; b) agents have no access to dedicated means of signaling. The study of unsignalized intersections is further motivated by the fact that they constitute part of the standards for crash avoidance research for decades \citep{Najm07}. 

\begin{figure*}
\centering
\includegraphics[width = .97\linewidth]{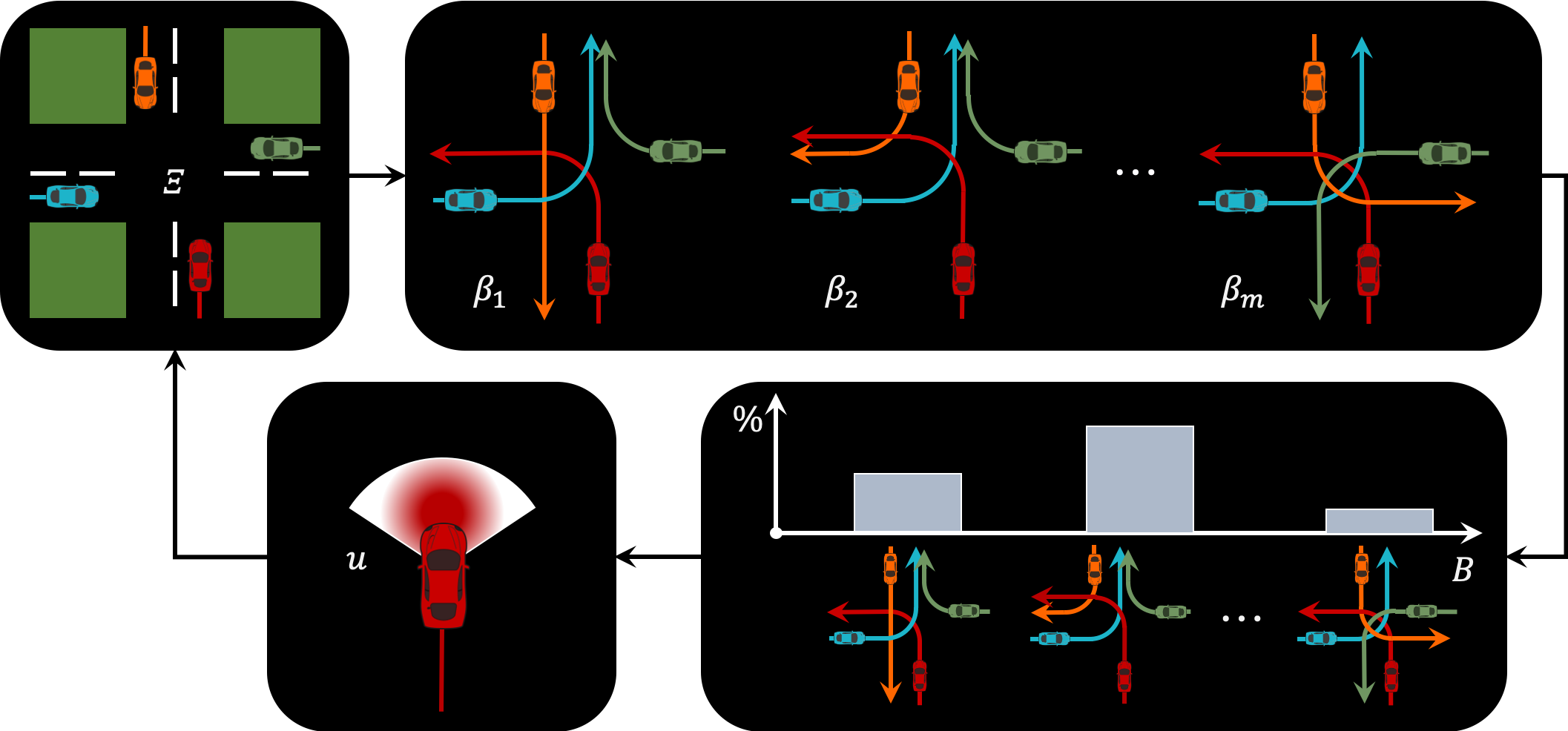}
\caption{Illustration of the decision-making scheme. At every cycle, the ego agent forward simulates a set of distinct futures, classifies them into topological outcomes, and selects the action that minimizes the uncertainty over such outcomes.\label{fig:flowchart}}
\end{figure*}

The lack of explicit communication in these domains results in high uncertainty about the unfolding dynamics (agents' intended destinations, trajectories, behavior models etc), which makes decision making challenging.  Intuitively, although cars can often be reasonably confined to remain in a particular segment of road or lane, the behavior inference problem is still challenging due to the need to reason on the continuum of possible intended trajectories.  Our key insight is that the spatial structure of the environment constrains the collective behavior of rational agents to belong to a finite set of \emph{discrete modes}, each corresponding to topologically distinct system behaviors.  We encode this domain knowledge into our approach by explicitly modeling these modes as topological \emph{braids} \citep{artin,birman}, removing the need for human guidance or learning such modes \textit{de novo} from examples. The braid formalism enables the abstraction of complex multiagent behaviors into topological modes, identified symbolically in a compact and interpretable fashion. This allows for the construction of a probabilistic inference mechanism that predicts future multiagent behaviors, represented as symbols, from observations of past multiagent trajectories. Based on this mechanism, a navigating agent may reason about the \emph{emerging mode}, i.e., the topological class capturing the emerging multiagent trajectory from the current time to the end of the execution. Understanding the likelihood and quality of the modes that are possible at a given configuration, agents can make principled decisions that balance personal efficiency with joint efficiency and safety. Based on this mechanism, we design a decentralized navigation planning framework that selects actions towards minimizing uncertainty over an emerging mode of behavior (see \figref{fig:flowchart}). By collectively contributing towards uncertainty reduction over the emerging mode through their action selection, the system of agents collectively reject unsafe executions. This dynamics is shown to result in safe executions despite the lack of explicit communication and signaling. 


In summary, within this work, we make the following contributions: 
\begin{enumerate}[label=(\alph*)]
    \item \textbf{We illustrate the virtues of topological braids for multiagent navigation in realistic environments.} We employ topological braids \citep{birman} as a formalism that may represent salient spatiotemporal features of multiagent navigation behavior in a compact and interpretable fashion. We build upon and extend past work on the use of braids for motion planning \citep{diazmercadobraidstro,mavrogiannis_ijrr} by providing a mathematically rigorous methodology for transitioning between Cartesian trajectories and topological braids and by considering a more realistic, obstacle-occupied structured domain (street intersection) with non-communicating nonholonomic agents (Sec.~\ref{sec:multimodality}).
    
    
    \item \textbf{We show how braids enable efficient inference by compressing the space of outcomes in structured environments.} Leveraging the spatial features of the ubiquitous structure of a street intersection, we construct a probabilistic inference mechanism that connects past agent trajectories to likely modes of future multiagent interaction (Sec.~\ref{sec:planning}). Reasoning about a bounded set of discrete, formally derived modes as opposed to trajectories has the potential of relaxing the prediction problem: under mild assumptions on agents' behavior (no U-turns and rationality), the space of modes is dramatically smaller than the corresponding space of trajectories.
    
    \item \textbf{We show that reasoning about braids results in safer executions.} We conduct an empirical study in which we compare our framework against a set of baselines that reason directly over the space of trajectories (Sec.~\ref{sec:application}). We demonstrate that our framework enables \emph{multiple} (2-4) non-communicating agents to coordinate implicitly and follow significantly safer paths (at least 65\% fewer collisions) across a series of challenging intersection-crossing scenarios. Our findings suggest that incorporating topological features in the decision making process of non-communicating agents enables effective coordination even in the absence of explicit communication.
    
    \item \textbf{We show how collectively individual actions may communicate complex multiagent intent.} Our findings illustrate the power of low-dimensional control actions in communicating complex multi-dimensional events, such as strategies of collision avoidance. We show that under the assumption of rationality on agents' decision making (no incentive of actively pursuing collisions, and goal-directedness), even simple prediction models could prove sufficient for collision avoidance, while exhibiting acceptably efficient behaviors.
\end{enumerate}

\section{Related Work}\label{sec:related-work}


Safe multi-agent planning in street intersections is a notoriously challenging problem, as is typically involves negotiation and coordination among multiple agents, often in the absence of explicit communication.  Developing autonomous systems capable of making safe decisions under such settings remains a challenge, leading to extensive research on the design of prediction, planning and control techniques.

Ensuring safety while maintaining efficiency is the key objective driving current research endeavors. \citet{Pierson_ICRA18} introduce a congestion cost quantifying the risk of collision and use it to plan within desired risk level sets for safe lane changes in congested highways. \citet{McGill19} present a probabilistic framework for automated crossing of unsignalized intersections under occlusions and faulty perception, which was shown to result in safe behaviors in real-world experiments on miniature racecars. \citet{Isele18} learn a policy for crossing unsignalized intersections under occlusions using deep reinforcement learning and show how it outperforms selected rule-based baselines. Finally, \citet{okamoto_ITS18} plan safe maneuvers at intersections by combining data-driven models for local and global vehicle interaction prediction.

A set of works model the problem of crossing an intersection using tools from belief-space planning. For instance, \citet{Bandyopadhyay_wafr13}, use a Mixed-Observability Markov Decision Process (MOMDP) to plan safe human intention-aware maneuvers in real-world vehicle-pedestrian interaction scenarios. Their approach has also been shown to enable safe merging in T-junction intersections \citep{Sezer_IROS15}. \citet{Bouton17} plan safe and efficient maneuvers for merging in unsignalized intersections using a partially observable Markov decision process model (POMDP) solved via a Monte Carlo sampling-based method. \citet{Hubmann_IV17} also propose a POMDP-based planner that incorporates uncertainty related to sensor noise besides intentions.

A class of works integrates a series of prosocial metrics on top of intention prediction towards reinforcing vehicle coordination. \citet{Sadigh2018} plan intent-expressive maneuvers that reinforce safe and efficient coordination between autonomous and human-driven cars at intersections and highways in a series of experiments on a driving simulator. Similarly, \citet{Lazar_CDC18} plan optimal lane changes that reinforce prosocial behaviors such as platooning, yielding increased capacity in congested highways. \citet{Buckman2019SharingIC} plan prosocial vehicle rearrangements that result in reduced system delays in a centrally managed signalized intersection, using a social psychology metric. Also within the centralized domain, \citet{MiculescuKaraman19} present a control framework inspired by polling systems that provides safety and efficiency guarantees for continuous car flows crossing an unsignalized intersection.

Other works have studied the problem of using discrete, semantic representations of traffic behaviors to predict the behavior of agents and their interactions with one another, and application to intent prediction and decision-making.  \citet{wang_understanding_2018} develop an approach to classifying discrete driving styles based on multi-dimensional time series analysis of a large corpus of data using a variant of hidden Markov models (HMM).  \citet{gadepally_framework_2017} use HMM to estimate long-term driver behaviors from a sequence of discrete decisions.  Others, such as \citet{konidaris_skills_2018} and \citet{shalev-shwartz_safe_2016}, propose using learned symbolic representations for high-level planning and collision avoidance, via a hierarchical options model.
While these works all serve to uncover a discrete representation of driving behaviors, they either require a large dataset to learn discrete modes or specified them by hand without harnessing the rich topological structure needed for collision-free planning.

Finally, a series of works have focused on developing tools for testing and validating approaches for autonomous navigation in realistic scenarios involving traffic at intersections. For instance, \citet{tian19} model traffic at unsignalized intersections using tools from game theory and propose a verification testbed for autonomous navigation algorithms. Similarly, \citet{Liebenwein20} propose a framework for safety verification of driving controllers based on compositional and contract-based principles, and validate it through a case study on a realistic road network. \citet{Gu_IATSS17} plan humanlike behaviors at intersections involving vehicle-pedestrian traffic using a data-driven model. Finally, \citet{hsu18}, also focusing on vehicle-pedestrian interactions at intersections explore how velocity signals generated by Markov decision processes affect interaction dynamics.





While existing literature focuses on the computational machinery for robust decision making under uncertainty, this paper identifies two key components that to the best of our knowledge have not been thoroughly studied in this domain: a) a salient mathematical representation that captures critical features of multiagent collision avoidance at intersection scenarios; b) a pipeline that leverages the implicit communication phenomena arising naturally while multiple agents navigate in a shared environment. Our insight is that effective incorporation of these features into the decision-making process of rational agents may enable efficient coordination despite the absence of explicit communication. 

In this paper, we formally model the structure of joint decision making at street intersections using the representation of topological braids \citep{birman}. Based on this representation, we build a probabilistic inference mechanism that predicts future joint behaviors of agents in the form of a probability distribution over topological braids. By selecting actions that are uncertainty-reducing over a space of future braids, non-communicating agents are able to converge to a collision-free protocol for intersection crossing. Overall, this paper is close in principle to the work of Mavrogiannis et al. \citep{MavKne_WAFR_2016,MavBluKne_IROS2017,MavKne_WAFR_2018,mavrogiannis_ijrr} but differs in that a) it considers a more rigorous mathematical introduction of braids and construction of braids from trajectories and b) it considers a more realistic, structured setup. Specifically, this paper features an analytical description of braids by adapting the presentation of \citet{wilson} to a multiagent trajectory using tools from calculus. This description constitutes a more rigorous but also a more accessible introduction to braids. Furthermore, unlike \citet{mavrogiannis_ijrr} who considered a discrete workspace and in contrast to \citet{MavBluKne_IROS2017} who considered holonomic agents in an obstacle-free workspace, in this paper, we focus on a continuous workspace with a realistic domain structure and nonholonomically-constrained agents following simple-car kinematics. 

The environment structure enables us to illustrate the power of topological braids even further that past work --it acts like a scaffold that bounds the motion of agents to the extent that the topology becomes the dominant property of agents' collision avoidance. This property yields beneficial computational implications as well as it allows us to cast the conventional explicit trajectory prediction problem as a symbolic topology prediction over a significantly smaller space of outcomes. Moreover, the structure provides geometric features that enable the design of a novel Bayesian inference mechanism extending the mechanism presented by \citet{mavrogiannis_ijrr}. Finally, unlike the work of \citet{diazmercadobraidstro} who employed braids to problems of centralized control, our framework focuses on the decentralized planning paradigm, explicitly tackling the implicit communication regime. Our work is inspired by recent works in human-robot collaboration which leverage the power of implicit communication to enhance robot performance in variety domains \citep{Tellex-RSS-14,DraganAuR14,knepper_hri_2017}.

\section{Problem Statement}\label{sec:statement}

\begin{figure}
\centering
\includegraphics[width = .97\linewidth]{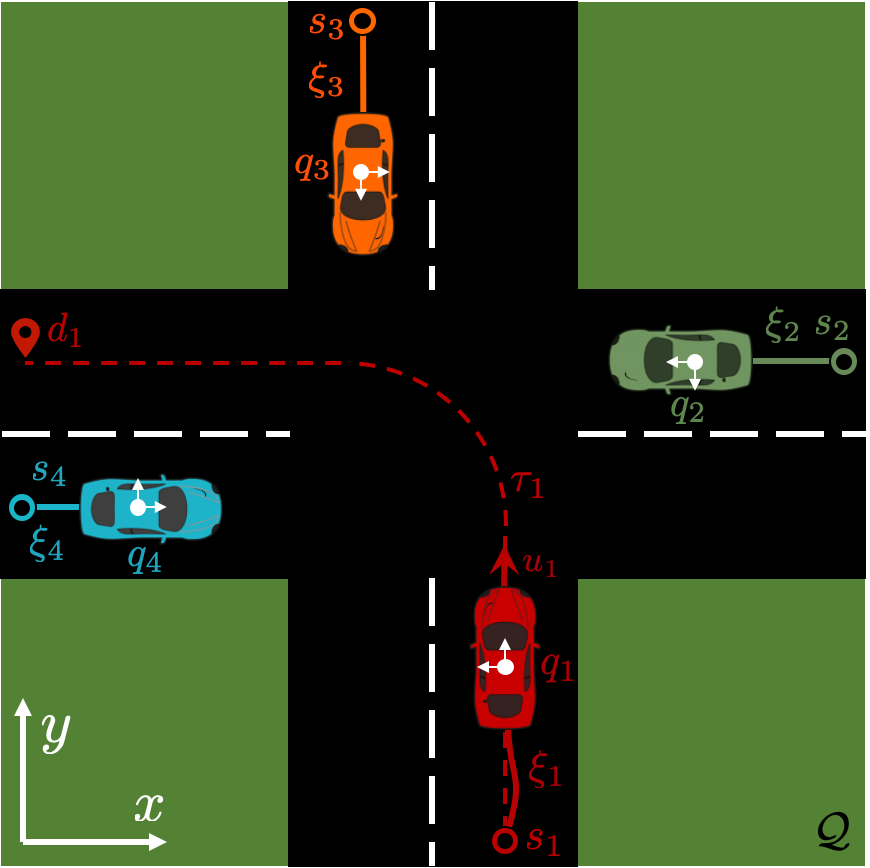}
\caption{Illustration of the notation. Having started from state $s_{1}$ and followed a trajectory $\xi_{1}$, the ego car (shown in red), currently in configuration $q_{1}$ and executing a control input $u_{1}$ follows a path $\tau_{1}$ towards its destination $d_{1}$. The figure also lists all quantities about other cars that are observable by the ego car. \label{fig:notation}}
\end{figure}

Consider the unsignalized street intersection of \figref{fig:notation} where $n\geq 1$ agents are navigating. Denote by $q_{i} = (x_{i},y_{i},\theta_{i})\in \pazocal{Q}\subseteq SE(2)$ the state of agent $i\in \pazocal{N} = \{1,\dots, n\}$ with respect to (wrt) a fixed reference frame, defined by a basis $(\hat{x},\hat{y},\hat{t})$. Agent $i$ is a dynamical system $\dot{q}_{i} = \phi(q_{i}, u_{i})$ following standard car dynamics \citep{Lav06}. Agent $i$ starts from an initial state $s_{i}\in \pazocal{Q}$, lying on a side of the intersection, and moves towards a final --unknown to others-- state $d_{i}\in\pazocal{Q}$ lying on a different side. Agent $i$ does so by tracking a path $\tau_{i}:I\to \pazocal{Q}$, for which it holds that $\tau_{i}(0) = s_{i}$ and $\tau_{i}(1) = d_{i}$, where $I=[0,1]$ is a path parametrization. Observing the complete system state $Q = (q_{1},\dots, q_{n})\in\pazocal{Q}^n$, agent $i$ tracks $\tau_{i}$ by executing a policy $\pi_{i}:\pazocal{Q}\to\pazocal{U}_{i}$, generating actions $u_{i}\in \pazocal{U}_{i}$ (speed and steering angle), satisfying a specification:
\begin{equation}
    u_{i} = \arg\min_{u_{i}\in\pazocal{U}_{i}} w_{i} C_{d}(u_{i}) + (1-w_{i})C_{c}(u_{i})\mbox{,}\label{eq:spec}
\end{equation} 
where $\pazocal{U}_{i}\subseteq\mathbb{R}\times\mathbb{S}$ is a space of controls, $C_{d}:\pazocal{U}_{i}\to\mathbb{R}_{\geq 0}$ represents the distance cost-to-go and $C_{c}:\pazocal{U}_{i}\to\mathbb{R}_{\geq 0}$ the collision cost of taking an action in consideration $u_{i}$, and $w_{i} \in (0, 1)$ is a weight --unknown to other agents-- describing agent $i$'s personal compromise over the two costs. Agent $i$ is not aware of the intended path $\tau_{j}$, the destination $d_{j}$ or the exact policy $\pi_{j}$ of agent $j\neq i\in \pazocal{N}$ but assumes that any agent $j\neq i\in \pazocal{N}$ is rational, in the sense that they also optimize for some compromise $w_{j}$ between $C_{d}$ and $C_{c}$. Our goal is to design decentralized policies $\pi_{i}$ that enable agents to coordinate safe intersection crossings while following time-efficient trajectories under uncertainty.

\section{Topological Multimodality of Intersection Crossing}\label{sec:multimodality}

The foundation of our approach lies in the observation that a constrained environment such as a street intersection couples the control decisions of rational agents. This coupling constrains collision-free multiagent trajectories to belong to a set of \emph{modes}, each corresponding to a distinct equivalence class of executions with identical topological properties. Our key insight is that explicitly reasoning about these modes during execution: (a) relaxes the inference problem, under the assumption that agents are acting rationally (i.e., they intend to follow short, collision-free paths); (b) enables agents to understand and represent potential solutions to the coupled collision-avoidance problem despite their uncertainty over the intentions or the policies of others. In this paper, we show that the modes of joint behavior at intersections can be modeled as topological braids \citep{artin, birman}. We then design an inference mechanism that predicts future braids given observations of past trajectories, and describe a policy generating uncertainty-minimizing actions to enable coordination among non-communicating agents.

\subsection{Joint Behavior at Street Intersections}

The complete sequence of controls that agent $i$ executes by tracking $\tau_{i}$ with the policy $\pi_{i}$, under the dynamics $\phi_{i}$, results in a time-parametrized trajectory $\xi_{i}:[0,t_{\infty}]\to\pazocal{Q}$, where $t_{\infty}$ corresponds to the end of the execution (the time at which the last agent reaches its destination --we assume without loss of generality that agents that reached their destinations earlier remain stationary until $t_{\infty}$). Following their individual policies, at time $t\in [0,t_{\infty}]$, the system of agents executes a control profile $U=\left(u_{1},u_{2},\dots, u_{n}\right)\in\mathbfcal{U}$, where $\mathbfcal{U}=\pazocal{U}_{i}\times \pazocal{U}_{2}\times \dots \times \pazocal{U}_{n}$ is the joint space of controls. Collectively, the complete sequence of control profiles that the system of agents executes from time $t=0$ to time $t=t_{\infty}$ to track the system path $T=\left(\tau_{1},\tau_{2},\dots,\tau_{n}\right)\in\pazocal{T}$ (where $\pazocal{T}$ represents the set of system paths) results in a time-parametrized system trajectory $\Xi = (\xi_{1},\dots, \xi_{n}): [0,t_{\infty}]\to \pazocal{Q}^n$. 


Depending on the relationship among the time parametrizations of agents' individual trajectories, the system trajectory $\Xi$ may exhibit different topological properties. These properties are indicative of the joint behavior of the system of agents, as they capture the succession with which agents traverse the intersection, e.g., which agent passed first/second, left/right (see \figref{fig:spatiotemporal}). We classify system trajectories into a set of modes, each corresponding to an equivalence class of topologically equivalent joint behaviors, represented as a topological braid \citep{artin,birman}.

\subsection{Topological Braids\label{sec:braids}}

Consider the intersection scenario depicted in \figref{fig:controldiagrams}. \figref{fig:pathsfollowed} depicts the trajectories followed by agents throughout the execution of that scenario, shown from a top view ($x$-$y$ projection), whereas \figref{fig:spatiotemporal} depicts the executed trajectories from a side view ($x$-$t$ projection). The latter projection unveils critical events that shaped the agents' joint collision-avoidance maneuvers through space and time. Following the theory of topological braids \citep{artin}, we can formally encode the succession of such events into symbols as shown in \figref{fig:spatiotemporal}. Braids are topological objects with geometric and algebraic descriptions, which enable us to abstract multiagent behaviors into symbolic motion primitives.

A braid is a tuple $(f_{1},\dots, f_{n})$ of functions $f_{i}:I\to\mathbb{R}^2\times I$, $i\in \pazocal{N}$, defined on a domain $I = [0,1]$ and embedded in a euclidean space $(\hat{x},\hat{y},\hat{t})$. These functions, called \textit{strands}, are monotonically increasing along the $\hat{t}$ direction and satisfy the following properties: (a) $f_{i}(0) = (i,0,0)$, and $f_{i}(1) = (p_{f}(i),0,1)$, where $p_{f}: \pazocal{N}\to \pazocal{N}$ is a permutation in the symmetric group $N_{n}$; (b) $f_{i}(t)\neq f_{j}(t)$ $\forall$ $t\in I$, $j\neq i\in \pazocal{N}$. The set of all braids on $n$ strands, along with the composition operation form a group, called the braid group on $n$ strands, denoted as $B_{n}$. The composition of two braids $b_{f} = (f_{1},\dots,f_{n})$, $b_{g} = (g_{1},\dots,g_{n})$, is also a braid $b_{h} = b_{f}\cdot b_{g} = (h_{1},\dots, h_{n})$, comprising a set of $n$ curves, defined as:
\begin{equation}
h_{i}(t) = \begin{dcases} f_{i}(2t), & t\in\left[0,\frac{1}{2}\right] \\ g_{j}(2t-1), & t\in\left[\frac{1}{2},1\right]\end{dcases}\mbox{,}
\end{equation}
where $j = p_{f}(i)$ ensures consistent indexing across $h$.

Following Artin's presentation \citep{artin}, the braid group $B_{n}$ can be generated from a set of $n-1$ primitive braids, $\sigma_{1},...,\sigma_{n-1}$, called generators (see \figref{fig:BnGenerators}), that satisfy the following \textit{relations}:
\begin{align}&\sigma_{i}\sigma_{j} =\sigma_{j}\sigma_{i},\qquad &|j-i|>1,\\ \qquad  &\sigma_{i}\sigma_{i+1}\sigma_{i}=\sigma_{i+1}\sigma_{i}\sigma_{i+1},\qquad &\forall\: i\mbox{.}\end{align}
\begin{figure}
\centering
\scalebox{.9}{
  \begin{tabular}{c c c}  
    \begin{minipage}[b]{0.3\linewidth}
    \centering
    \begin{tikzpicture}
      \braid[number of strands=5, line width=2pt, width = 12pt, height = 55pt,  style strands={1}{black},
 style strands={2}{red},
 style strands={3}{black},  style strands={4}{black}, style strands={5}{black}]
      a_1^{-1} a;
      \node[font=\Huge] at (1.5,-1.3) { \(\mydots\) };
    \end{tikzpicture}\\
    \small{(a) $\sigma_{1}$}
    \end{minipage}    
        
    \begin{minipage}[b]{0.3\linewidth}
    \centering
          \begin{tikzpicture}
      \braid[number of strands=5, line width=2pt, width = 12pt, height = 55pt,  style strands={1}{black},
 style strands={2}{black},
 style strands={3}{red},  style strands={4}{black},  style strands={5}{black}]
      a_2^{-1} a;
      \node[font=\Huge] at (1.9,-1.3) { \(\mydots\) };
    \end{tikzpicture}\\
          \small{(b) $\sigma_{2}$}
    \end{minipage}    
    
    \begin{minipage}[b]{0.1\linewidth}
    \centering
    \raisebox{3.2em}{
          \begin{tikzpicture}
      \node[font=\large] { \(\cdots\) };
    \end{tikzpicture}}\\
    \end{minipage}
    
    \begin{minipage}[b]{0.3\linewidth}
    \centering
          \begin{tikzpicture}
      \braid[number of strands=5, line width=2pt, width = 12pt,  height = 55pt,  style strands={1}{black},
 style strands={2}{black},
 style strands={3}{black},  style strands={4}{black},  style strands={5}{red}]
      a_4^{-1} a;
      \node[font=\Huge] at (1.05,-1.3) { \(\mydots\) };
    \end{tikzpicture}\\
          \small{(c) $\sigma_{n-1}$}
    \end{minipage}\\  
     & &  
  \end{tabular}
  }
  \vspace{-.5cm}
\caption{\small{The generators of the Braid Group $B_{n}$.}\label{fig:BnGenerators}}
\end{figure}
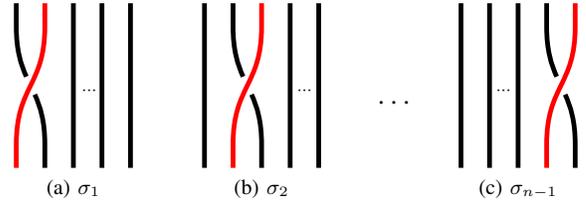
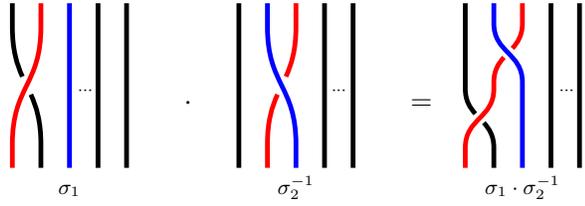
\begin{figure}
\centering
\scalebox{.9}{
  \begin{tabular}{c c c}  
    \begin{minipage}[b]{0.3\linewidth}
    \centering
    \begin{tikzpicture}
      \braid[number of strands=5, line width=2pt, width = 12pt, height = 55pt, style strands={1}{black},
 style strands={2}{red},
 style strands={3}{blue},  style strands={4}{black},  style strands={5}{black}]
      a_1^{-1} a;
      \node[font=\Huge] at (1.5,-1.3) { \(\mydots\) };
    \end{tikzpicture}\\
    \small{$\sigma_{1}$}
    \end{minipage}    
    
    \begin{minipage}[b]{0.05\linewidth}
    \centering
    \raisebox{3.2em}{
          \begin{tikzpicture}
      \node[font=\large] { \(\cdot\) };
    \end{tikzpicture}}\\
    \end{minipage}
        
    \begin{minipage}[b]{0.3\linewidth}
    \centering
          \begin{tikzpicture}
      \braid[number of strands=5, line width=2pt, width = 12pt, height = 55pt, style strands={1}{black},
 style strands={2}{blue},
 style strands={3}{red},  style strands={4}{black},  style strands={5}{black}]
      a_2 a;
      \node[font=\Huge] at (1.9,-1.3) { \(\mydots\) };
    \end{tikzpicture}\\
          \small{$\sigma_{2}^{-1}$}
    \end{minipage}    
    
    \begin{minipage}[b]{0.05\linewidth}
    \centering
    \raisebox{3.2em}{
          \begin{tikzpicture}{}
      \node[font=\large] { \(=\) };
    \end{tikzpicture}}\\
    \end{minipage}
    
    \begin{minipage}[b]{0.3\linewidth}
    \centering
          \begin{tikzpicture}
      \braid[number of strands=5, line width=2pt, width = 12pt,  height = 27.5pt, style strands={1}{black},
 style strands={2}{blue},
 style strands={3}{red},  style strands={4}{black},  style strands={5}{black}]
      a_2 a_1^{-1}  ;
      \node[font=\Huge] at (1.92,-1.3) { \(\mydots\) };
    \end{tikzpicture}\\
          \small{$\sigma_{1}\cdot\sigma_{2}^{-1}$}
    \end{minipage}\\  
     & &  
  \end{tabular}
  }
  \vspace{-.5cm}
\caption{\small{The Composition $\sigma_{1}\cdot\sigma_{2}^{-1}$ for the Braid Group $B_{n}$.}\label{fig:composition}}
\end{figure}
%
A generator $\sigma_{i}$ is a braid $\sigma_{i} = (\upsigma_{1},\dots,\upsigma_{n})$, $i\in \pazocal{N}\backslash n$ with the following properties: (a) $\sigma_{i}(0) = (1,0,0)$, and $\sigma_{i}(1) = (p_{i}(i),0,1)$, where $p_{i}: \pazocal{N}\to \pazocal{N}$ is an adjacent transposition (a permutation swapping exactly two adjacent elements) swapping the elements $i$ and $i+1$; (b) there exists a unique $t_{c}\in [0,1]$ such that $(\sigma_{i}(t_{c})-\sigma_{i+1}(t_{c}))\cdot \hat{x} = 0$ and also $(\upsigma_{i}(t_{c})-\upsigma_{i+1}(t_{c}))\cdot \hat{y} >0$.

Similarly, the inverse of $\sigma_{i}$ is the braid $\sigma^{-1}_{i}= (\upsigma^{-1}_{1},\dots,\upsigma^{-1}_{n})$, $i\in \pazocal{N}\backslash n$ with the following properties: (a) $\sigma^{-1}_{i}(0) = (1,0,0)$, and $\sigma^{-1}_{i}(1) = (p_{i}(i),0,1)$; (b) there exists a unique $t_{c}\in [0,1]$ such that $(\sigma^{-1}_{i}(t_{c})-\sigma^{-1}_{i+1}(t_{c}))\cdot \hat{x} = 0$ and also $(\upsigma_{i}(t_{c})-\upsigma_{i+1}(t_{c}))\cdot \hat{y} < 0$.

Finally, the identity braid $\sigma_{0} = (\sigma^{0}_{1},\dots, \sigma^{0}_{n})$ is defined via a trivial permutation implementing no swap, i.e., $p_{0}(i) = i$ and thus yielding $\upsigma^{0}_{i}(0) = (i,0,0) = (i,0,1)$ while it also holds that $\nexists t_{c}\in [0,1]$ such that $(\sigma^{0}_{i}(t_{c})-\sigma^{0}_{i+1}(t_{c}))\cdot \hat{x} = 0$ for any $i$.

Any braid can be written as a product of generators and generator inverses. This product is commonly referred to as a \emph{braid word}. \figref{fig:composition} depicts an example of a braid product (composition of two braids) whereas \figref{fig:BnGenerators} depicts the braid group on $n$ strands $B_{n}$.

\subsection{Multiagent Intersection Crossing as a Topological Braid}\label{sec:trajectoriesbraids}

Consider the tuple $\Xi = (\xi_{1},\dots,\xi_{n})$ containing the trajectories of $n$ agents as they traverse the intersection of \figref{fig:notation}. We will now show that starting from $\Xi$ and by following a sequence of topology-preserving operations, we can construct a braid $\beta = (\xi^{+}_{1},\dots, \xi^{+}_{n})$ with the same topological properties as $\Xi$. This will enable us to employ braids as symbols corresponding to multiagent motion primitives, based on which we will design an inference mechanism.

Define by $\xi^{x}_{i}:[0,t_{\infty}]\to\mathbb{R}$ and $\xi^{y}_{i}:[0,t_{\infty}]\to\mathbb{R}$ the $x$ and $y$ coordinates of the trajectory of agent $i$. For $t=0$, ranking agents in order of increasing $\xi^{x}_{i}(0)$ value defines a starting permutation $p_{s}:\pazocal{N}\to \pazocal{N}$, where $p_{s}(i)$ denotes the rank of agent $i$. For $t=t_{\infty}$, ranking agents in order of increasing $\xi^{x}_{i}(t_{\infty})$ value defines a final permutation $p_{d}:\pazocal{N}\to \pazocal{N}$, where $p_{d}(p_{s}(i))$ denotes the final ranking of agent $i$. Further, define $x_{min} = \min_{i,t} \xi^{x}_{i}(t)$, $x_{max} = \max_{i,t} \xi^{x}_{i}(t)$, and $y_{min} = \min_{i,t} \xi^{y}_{i}(t)$, $y_{max} = \max_{i,t} \xi^{y}_{i}(t)$. Assuming that $x_{max}\neq x_{min}$ and $y_{max}\neq y_{min}$, which is compatible with the intersection domain considered in the paper, we define the ratio functions
\begin{equation}
    r_{i}^{x}(t) = \frac{\xi^{x}_{i}(t)-x_{min}}{x_{max} - x_{min}}
\end{equation}
and 
\begin{equation}
    r_{i}^{y}(t) = \frac{\xi^{y}_{i}(t)-y_{min}}{y_{max} - y_{min}}\mbox{,}
\end{equation}
which allow us to keep track of the relationships among the $x$ and $y$ coordinates of agents in a normalized way. Note that

Denote by $\uptau:I\to [0, t_{\infty}]$ a time parametrization function, uniformly mapping the domain $I=[0,1]$ to the execution time in the range $[0,t_{\infty}]$. We define a set of possibly-discontinuous functions $\left(\xi_{1}^{+},\dots,\xi^{+}_{n}\right)$, with $ \xi^{+}_{i}:I\to\mathbb{R}^2\times I$, $i\in \pazocal{N}$ such that:
\begin{equation}
\begin{split}
& \xi^{+}_{p_{s}(i)}(a) =\\
& \quad\begin{dcases}
\left(p_{s}(i),0,0\right), & a = 0\\
\left(1+r_{i}^{x}(\uptau(a))(n-1), -1 + 2r_{i}^{y}(\uptau(a)), a\right), & a \in \left(0,1\right)\\
\left(p_{d}(p_{s}(i)),0,1\right), & a = 1\\
\end{dcases}
\end{split}
\mbox{.}
\label{eq:xiplus}
\end{equation}

\begin{figure}
    \centering
    \begin{subfigure}{0.48\linewidth}
        \centering
        \includegraphics[trim = {0cm 0cm 0cm 0cm}, clip, width = \linewidth]{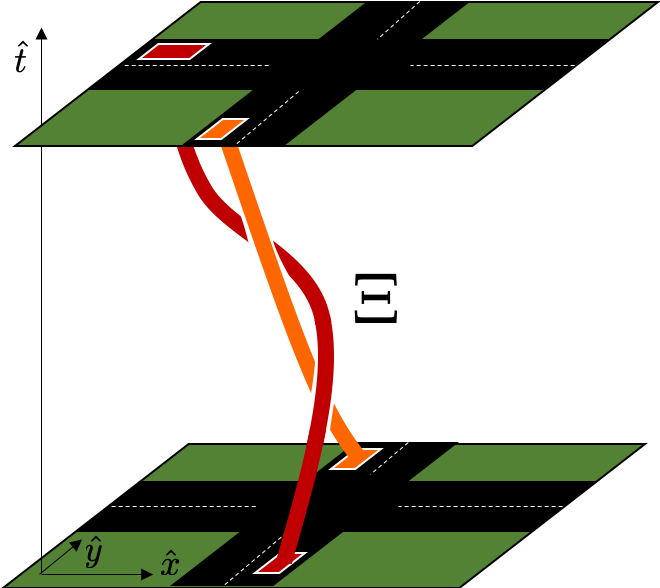}
        \caption{Trajectories of two agents as they cross an intersection, plotted in spacetime.\label{fig:intersection-perspective2}}
    \end{subfigure}
    ~    
    \begin{subfigure}{0.48\linewidth}
        \centering
        \includegraphics[trim = {0cm 0cm 0cm 0cm}, clip, width = \linewidth]{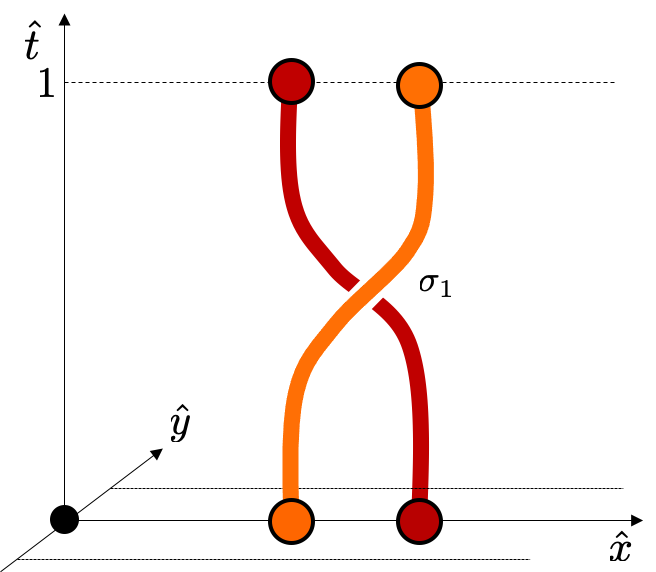}
        \caption{Braid $\sigma_{1}$ capturing the topological entanglement of  agents' trajectories.\label{fig:intersection-braid2}}
    \end{subfigure}
    ~    
    \begin{subfigure}{0.48\linewidth}
        \centering
        \includegraphics[trim = {0cm 0cm 0cm 0cm}, clip, width = \linewidth]{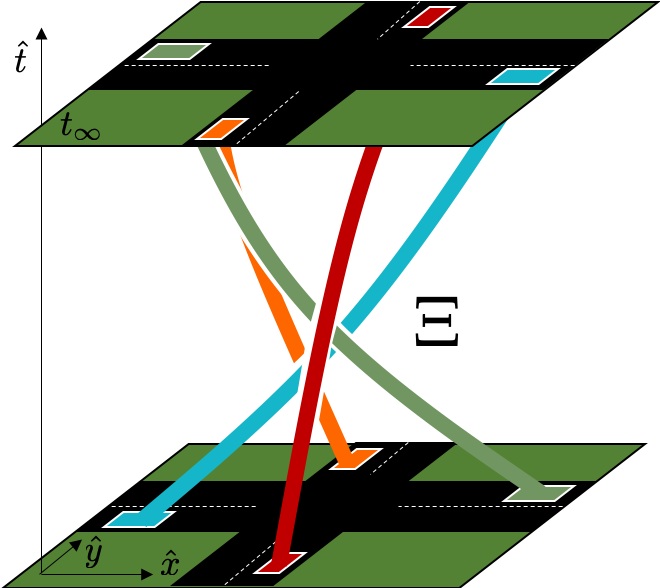}
        \caption{Trajectories of four agents as they navigate an intersection, plotted in spacetime.\label{fig:intersection-perspective}}
    \end{subfigure}
    ~
    \begin{subfigure}{0.48\linewidth}
        \centering
        \includegraphics[trim = {0cm 0cm 0cm 0cm}, clip, width = \linewidth]{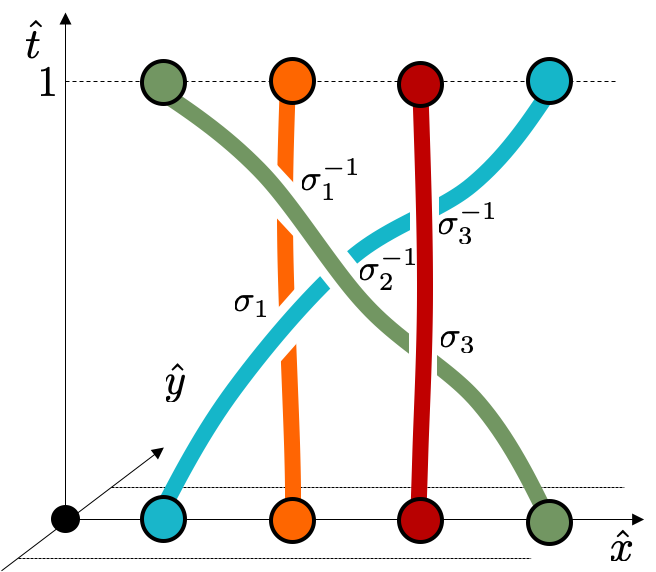}
        \caption{Braid $\sigma_{3}\sigma_{1}\sigma_{2}^{-1}\sigma^{-1}_{3}\sigma_{1}^{-1}$ capturing the topological entanglement of agents' trajectories.\label{fig:intersection-braid}}
    \end{subfigure}    
    \caption{Transition from an intersection crossing represented as a collection of Cartesian trajectories (Figs \subref{fig:intersection-perspective2}, \subref{fig:intersection-perspective}) to a topologically equivalent symbolic representation based on the formalism of topological braids (Figs \subref{fig:intersection-braid2}, \subref{fig:intersection-braid}) for scenarios involving two (top row) and four agents (bottom row) via eq.~\eqref{eq:xiplus}.}
    \label{fig:intersection2braid}
\end{figure}

The set of functions $\left(\xi_{1}^{+},\dots, \xi_{n}^{+}\right)$ forms a topological \emph{braid} $\beta$ following the definition of Sec.~\ref{sec:braids}. The time mapping $\uptau$ reparametrizes the domain of the trajectories to be within the range $[0,1]$. The endpoints of $\beta$ correspond to the permutations defined by the arrangement of agents in order of increasing $x$-projection. For $a\in (0,1)$, the expressions of \eqref{eq:xiplus} scale the $x$-coordinates to lie within the range $[1,6]$ and the $y$-coordinates to lie within the range $[-1,1]$ in a way that retains the relationships among them and thus preserves the topological properties formed within $\Xi$. Therefore, the braid $\beta$ is topologically equivalent (formally ambient-isotopic \citep{murasugi1999study}) to the execution $\Xi$. \figref{fig:intersection2braid} depicts an illustrated example of transitioning from a set of trajectories to a topological braid.

The process described above allows us to analyze the topological properties of $\Xi$ using tools from the theory of topological braids \citep{birman}. In particular, it allows us to abstract an execution represented as a multiagent trajectory $\Xi$ into a \emph{word} $\beta$ composed of topological symbols. We do so by following the approach of \citet{Thiffeault2010}: a) we first label the pairwise trajectory crossings that emerge within the $x$-$t$ projection as braid generators by identifying  \emph{under} or \emph{over} crossings (see \figref{fig:intersection-braid2}, \figref{fig:intersection-braid}) and b) we then arrange these generators in temporal order. The outcome of this process is a \emph{braid word} serving as a symbolic abstraction of an execution.

\begin{remark}
Note that the construction of a topological braid $\beta$ from a system trajectory $\Xi$ described above assumed the projection plane $\hat{x}$-$\hat{t}$ for convenience. Alternative spatiotemporal planes could be used, defined by a spatial vector $\hat{\eta}$ of one's choice and the time vector $\hat{t}$. Any system trajectory could be abstracted into a braid word, under the assumption of any vector $\hat{\eta}$. Any topological outcome could be captured and represented through the use of the method described above. However, in this work, we are interested in expressing topological outcomes of interest to the navigating agent. For this reason, in the remainder of the paper, we assume that every agent $i$, assumes the projection plane defined by the vector $\hat{x}_{i}$ (and time) as shown in \figref{fig:pathsets}. This projection ensures that for agent $i$, the braid formalism consistently captures critical spatial relationships for safely crossing the intersection. Intuitively, this projection enables the ego agent to distinctly represent topological outcomes corresponding to agents crossing or ``cutting" in front of it, thus allowing for proper inference and response to them.
\end{remark}

\subsection{Representing Distinct Intersection Crossings as Braids}

Besides describing the topological properties of a given multiagent trajectory, braids may be used as representatives of classes of topologically equivalent multiagent behaviors. In particular, at an intersection with $n$ agents, the braid group, $B_{n}$, can be thought of as the complete set of possible ways in which agents' trajectories could entangle with each other over time as they navigate between their initial and final configurations. 

\begin{figure}
    \centering
    \includegraphics[width = \linewidth]{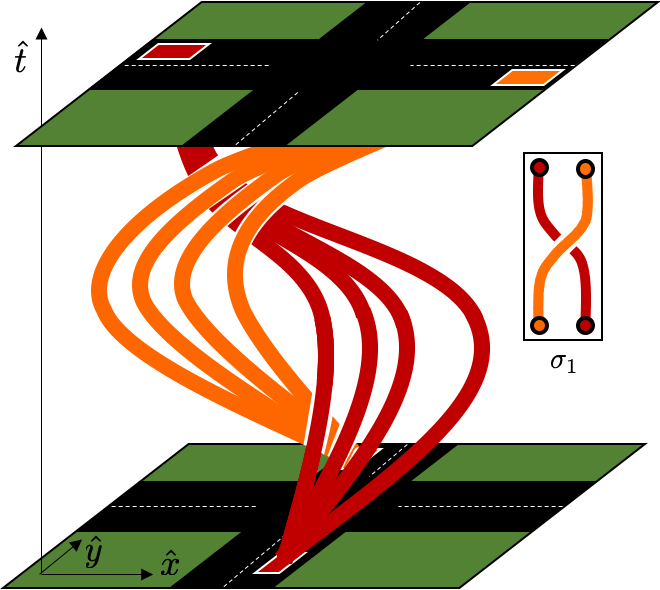}
    \caption{A braid serves as a representative of all topologically equivalent multiagent trajectories. In this example, fixing agents' start and goal locations, the braid $\sigma_{1}$ is a representative of all trajectory pairs in which the red agent turns to the left \emph{after} the orange agent enters the intersection.}
    \label{fig:braidisarepresentative}
\end{figure}

For instance, consider the scenario of \figref{fig:braidisarepresentative} in which the red agent moves from the bottom to the left side and the orange agent from the top to the right side of the intersection. Within that scenario, the braid $\sigma_{1}$ (defined assuming the projection $x$-$t$) constitutes a representative of all trajectory pairs in which the red agent turns to the left \emph{after} the orange agent enters the intersection. The group $B_{2}$ contains all ways in which the two agents could theoretically mix in that scenario, containing arbitrarily more complex braids, as described in Sec.~\ref{sec:braids}. However, under the assumption that agents are acting rationally (agents aiming for efficient and collision-free paths), only one more braid would be possible: $\sigma_{1}^{-1}$, in which the red agent turns to the left \emph{before} the orange agent enters the intersection.

This example highlights the connections among the topology of a multiagent trajectory, agents' destinations and agents' behavior mechanisms. In general, the likelihood of braids in $B_{n}$ is related to the shape of uncertainty about other agents' destinations and behavioral mechanisms but also on their observed state history so far. Based on this insight, Sec.~\ref{sec:inference} describes a simple inference mechanism, iterating across possible destinations and behaviors for other agents. This inference will be at the core of a decision making scheme, presented in Sec.~\ref{sec:decisionmaking}.

\section{Planning with Multimodal Inference Enabled by Topological Braids}\label{sec:planning}

\begin{figure}
\centering
\includegraphics[width = \linewidth]{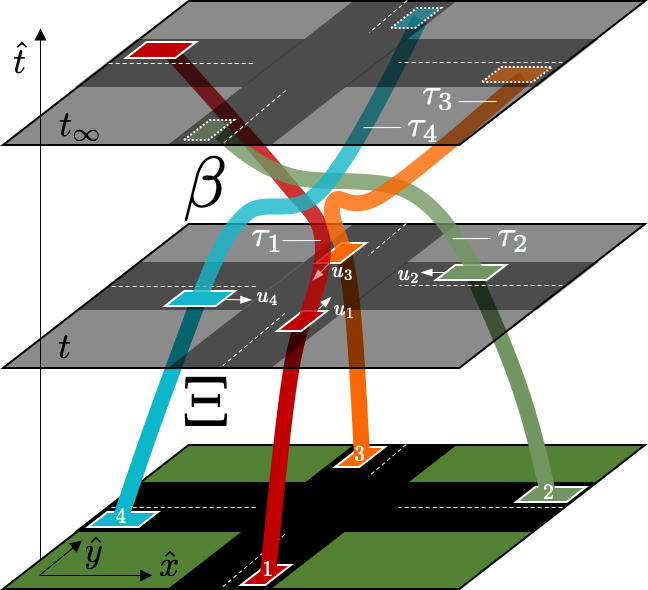}
\caption{Illustration of the inference architecture at a four-agent intersection crossing scenario. At time $t$, the ego agent (red color), heading to destination $\tau_{1}$, having access to the joint state history $\Xi$, updates its belief about the topology $\beta$ of the emerging multiagent interaction based on beliefs over other agents' new speeds $u_{i}$ and destinations $\tau_{i}$, $i\in\{2,3,4\}$.\label{fig:inference}}
\end{figure}

In this section, we describe a probabilistic model that links past agents' trajectories to a braid representing the spatiotemporal entanglement of their future trajectories at an intersection domain. \figref{fig:inference} illustrates the setup of the proposed model at a four-agent scenario. Based on this mechanism, we build a decision-making scheme for decentralized motion planning at unsignalized intersections.

\subsection{Multimodal Inference with Topological Braids\label{sec:inference}}

At time $t\in [0,t_{\infty}]$, agent $i$, having access to the complete system state history so far, $\Xi$, maintains a belief $bel_{i}(\beta_{i})=P(\beta_{i}|\Xi)$ over the braid $\beta_{i}\in B_{n}$ that describes the topology of the emerging (future) system trajectory $\Xi' = \Xi_{t\to \infty}$. This braid is extracted upon assuming a projection onto a plane defined by a vector $\hat{\eta}_{i}$ and time $\hat{t}$. For instance, in \figref{fig:braidisarepresentative}, the red agent (agent 1) has set $\hat{\eta}_{1}=\hat{x}$, assuming the projection plane $\hat{x}$-$\hat{t}$.

The braid $\beta_{i}$ is heavily dependent on agents' intended system path $T$. To capture this dependency, we marginalize over $\pazocal{T}_{i}\subseteq\pazocal{T}$, the subset of all system paths for which agent $i$ (the ego agent) follows its intended path:

\begin{equation}
bel_{i}= P(\beta_{i}|\Xi) = \sum_{T\in\pazocal{T}_{i}}P(\beta_{i}|\Xi,T)P(T|\Xi)\mbox{.}\label{eq:bel}
\end{equation}

For a given system path $T$, different braids could possibly emerge, depending on the path tracking behavior of agents. To capture this dependency, we marginalize the probability $P(\beta_{i}|\Xi,T)$ over the set of possible control profiles that could be taken by agents at the current time step:
\begin{equation}
P(\beta_{i}|\Xi,T)=\sum_{U\in\mathbfcal{U}}P(\beta_{i}|\Xi,U,T)P(U|\Xi,T)\mbox{.}\label{eq:bel2}
\end{equation}
Substituting to eq. \eqref{eq:bel}, we get:
\begin{equation}
bel_{i} = \sum_{\pazocal{T}_{i}} \Bigg\{\sum_{\mathbfcal{U}}P(\beta_{i}|\Xi,U,T)P(U|\Xi,T)\Bigg\}P(T|\Xi)\mbox{.}
\label{eq:finalbelief}
\end{equation}
The outlined mechanism combines a local action selection model $P(U|\Xi,T)$ with a model of intent inference $P(T|\Xi)$ and a global behavior prediction model $P(\beta_{i}|\Xi,U,T)$.

The intention of agent $j\neq i$ over a path $\tau_{j}$ is conditionally independent of the intention of any other agent, given the past system trajectory $\Xi$. The probability over the path intention of agent $j$ does not depend on the trajectories of others. Thus, we simplify the computation of the system path prediction as:
\begin{equation}
P(T|\Xi) = \prod_{j\in \pazocal{N} \backslash i}P(\tau_{j}|\xi_{j})\mbox{,}
\end{equation}
where the product only considers the probabilities over the paths of others, since agent $i$ is certain about its own path.

Similarly, since agents select a control input independently, without having access to the policies of others, we decompose the computation of the control profile prediction as:
\begin{equation}
P(U|\Xi,T) = \prod_{i=1}^{n}P(u_{i}|\Xi,T)\mbox{,}\label{eq:transitionmodel}
\end{equation}
where the distribution $P(u_{i}|\Xi,T)$ represents the control input that agent $i$ executes to make progress along its path $\tau_{i}$, incorporating considerations such as preferred navigation velocity and a local controller class.

The model of inference of eq.~\eqref{eq:finalbelief} focuses on topology prediction, without considerations of collision avoidance. To filter out unsafe braids, we redefine eq. \eqref{eq:finalbelief} by incorporating a model of collision prediction. Define by $c$ a boolean random variable representing the event that $\Xi'$, the emerging future trajectory contains collisions (\texttt{true} for a collision, \texttt{false} for no-collision). Denote by $\tilde{\beta} = (\beta_{i}, \neg c)$ the joint event that $\Xi'$ is both topologically equivalent (ambient-isotopic \citep{murasugi1999study}) to a braid $\beta_{i}\in B_{n}$, \emph{and} not in collision, i.e., $c$ is \texttt{false}. Then the probability that $\tilde{\beta_{i}}$ is true can be computed as:
\begin{equation}
bel_{i}(\tilde{\beta}_{i}) = \sum_{\pazocal{T}} \Bigg\{\sum_{U}P(\tilde{{\beta}}_{i}|\Xi,U,T)P(U|\Xi,T)\Bigg\}P(T|\Xi)\mbox{.}\label{eq:finalfinalbelief}
\end{equation}
The occurrence of a collision is conditionally independent of the emerging braid given the state history, the current control profile and the intended system path; thus, we may compute their joint distribution as:
\begin{equation}
\begin{split}
P(\tilde{\beta}_{i}|\Xi,U,T)=&P(\tilde{\beta}_{i},\neg c|\Xi,U,T)\\ =& P(\neg c|\Xi,U,T,\beta_{i})P(\beta_{i}|\Xi,U,T)\\ =& (1-P(c|\Xi,U,T,\beta_{i}))P(\beta_{i}|\Xi,U,T)\label{eq:finalfinal}
\mbox{.}
\end{split}
\end{equation}

\subsection{Decision Making}\label{sec:decisionmaking}


An outcome $\tilde{\beta}_{i}$ represents a class of solutions to the problem of multiagent collision avoidance at the intersection with distinct topological properties. Before execution, multiple classes of solutions could be theoretically possible. However, as execution progresses, the distribution over those classes, $P(\tilde{\beta}_{i}|\Xi,U,T)$, from the perspective of the ego agent (agent $i$) is reshaped as a result of agents' decisions. Our approach is based on the insight that by manipulating agents' actions over time we could enable them to reach a state of clear consensus over a solution $\tilde{\beta}_{i}$ more quickly. To this end, we employ a decision-making mechanism that contributes uncertainty-reducing actions over the emerging solution $\tilde{\beta}_{i}$ by directly minimizing the information entropy of the distribution over future braids(eq.~\eqref{eq:finalfinalbelief}). The lower the Entropy is, the lower the uncertainty, and thus the closer agents are to a consensus over a solution $\tilde{\beta}_{i}$. Besides, the use of the information entropy cost reflects the realization that multiple solutions from $B_{n}$ could be valid solutions to the collision avoidance problem at a given instance. The ego agent behavior could still contribute to collision-free navigation even when there is not a unique winner within $B_{n}$. This decision is motivated and in line with recent work in shared autonomy (e.g., \citet{javdanietal18}).

From the perspective of agent $i$, the uncertainty over a solution $\tilde{\beta}_{i}$ is represented as:
\begin{equation}
H(\tilde{\beta}_{i})=-\sum_{B_{n}}P(\tilde{\beta_{i}}|\Xi)\log P(\tilde{\beta_{i}}|\Xi)\mbox{,}
\end{equation}
where $P(\tilde{\beta_{i}}|\Xi)$ is recovered using eq. \eqref{eq:finalfinalbelief}. In order to contribute towards reducing this uncertainty, agent $i$ selects actions (velocities) that minimize the entropy:
\begin{equation}{}
u_{i}=\arg\min_{u_{i}}H(\tilde{\beta}_{i})\mbox{.}\label{eq:planning}
\end{equation}
Notice that eq.~\eqref{eq:planning} contains the tuple $U$ through the marginalization of eq.~\eqref{eq:finalfinalbelief}. Thus, the optimization scheme outputs the control input $u_{i}$ for agent $i$ that reduces $H(\beta_{i})$ the most in expectation.

\section{Application}\label{sec:application}

\begin{figure}
    \centering
    \includegraphics[width = \linewidth]{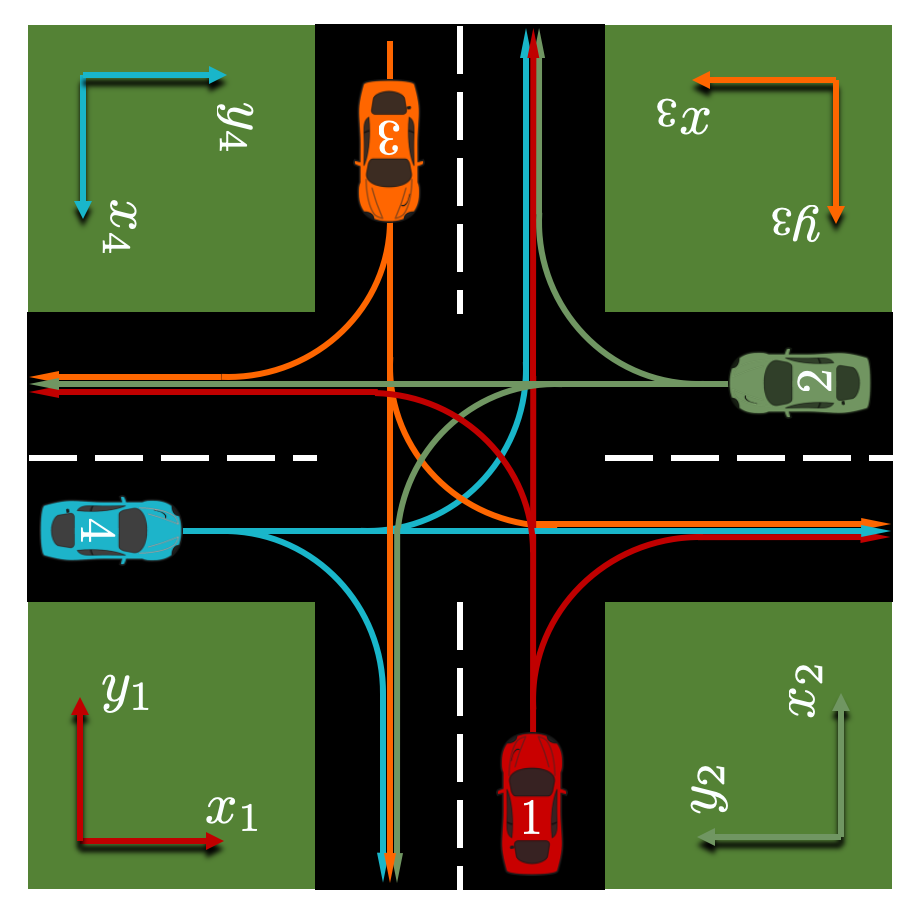}
    \caption{Experimental Setup. Agents' path sets and reference frames are drawn in the same color. Each agent selects a path from its path set (left, right, or forward) and represents braids by assuming a projection plane defined by $x_{i}$ and time (out of the page).}
    \label{fig:pathsets}
\end{figure}

We employ our decision-making mechanism in a simulated study on an unsignalized intersection with multiple vehicles. Our setup is the 4-way symmetric intersection of \figref{fig:notation}. Each lane at the intersection is $50m$ long and $3.6m$ wide, whereas each car is represented as a rectangle with a length of $4.7m$ and a width of $1.7m$. We assume that any side $a$ is connected to any different side $b\neq a$ with a unique, publicly known legal path $\tau_{ab}$, lying along the middle of the lane (see \figref{fig:pathsets}). We assume that any agent $i\in \pazocal{N}$ that attempts to reach side $b$ from side $a$ will attempt to track this path, $\tau_{ab}$. All agents track their paths with the same tracking controller $\kappa:V_{i}\to\pazocal{U}_{i}$, taking as input a desired linear velocity $v_{i}\in V_{i}$ in Cartesian coordinates and outputting a control command $u_{i}\in\pazocal{U}_{i}$. This controller is designed based on a feedback linearization technique \citep{Yun92onfeedback} although any other path tracking controller such as e.g., PID or pure pursuit \citep{Coulter-1992-13338} could be used. The main decision variable of our scheme is the speed with which an agent tracks its path; given that speed, the tracking controller outputs a control input $u_{i}\in\pazocal{U}_{i}$ which is directly executed.


Each agent follows a path out of three options (left, right, or straight) as shown in \figref{fig:pathsets}. Thus, agent $i$ needs to consider $|\pazocal{T}_{i}|=3^3=27$ possible system paths, extracted upon iterating over all possible combinations for other agents' paths. We consider path tracking to be split in two parts: (a) the \textit{negotiation} part, which corresponds to the initial straight-path part of the intersection (denoted as $\pazocal{Q}_{i}^{neg}$ for agent $i$), within which the agent attempts to reach a consensus with others wrt a joint strategy of collision avoidance; (b) the \textit{execution} part, which corresponds to the rest of the path (denoted as $\pazocal{Q}_{i}^{exec}$ for agent $i$), within which the agent tracks the remainder of its path, by maintaining a constant speed. This decision emphasizes the importance of proactive negotiation during the first portion, and provides a natural metric of quality --the count of collisions during the execution part. 


\subsection{Models}

We assume that agent $i$ has no knowledge of the path $\tau_{j}$ of any other agent $j\neq i\in \pazocal{N}$ while $j$ is in the negotiation stage. However, we assume that $\tau_{j}$ becomes immediately obvious when agent $j$ enters the intersection:
\begin{equation}
P(\tau_{j}|\xi_{j}(t)) = P(\tau_{j}|q_{j}) = 
\begin{cases}
1/m & \textnormal{for}\: q_{j}\in Q^{neg}_{j}\\
1 & \textnormal{for}\: q_{j}\in Q^{exec}_{j}\mbox{,}\\
\end{cases}
\end{equation}
where $q_{j} = \xi_{j}(t)$ is agent $j$'s current state, and $m=3$ is the number of paths that agent $j$ selects from.


At time $t$, agent $i$ selects a velocity $v_{i}\in V_{i}^{t}$ and passes it to the controller $\kappa$ which converts it to an executable control command $u_{i}$. The set $V_{i}^{t}$ contains two linear Cartesian velocities pointing towards the next waypoint in $\tau_{i}$: one of low magnitude $|v^{t}_{low}|$ and one of high magnitude $|v_{high}^{t}|$. These magnitudes are fixed throughout the execution and are generally distinct for each agent. We further assume that all agents prefer the higher speed, and that in the beginning of the execution, they start with the high speed. We express this preference in the probability $P(u_{j}|\xi_{j},\tau_{j})$. In the following simulations we assume that agents prefer the high speed with higher probability over the low speed. For each agent, this probability is sampled uniformly from the range $[0.6,0.8]$ and remains fixed throughout the execution. We also assume that each agent assumes that others have the same exact preferences over speeds, i.e., they do \emph{not} know the true preferences of others. This decision introduces a level of uncertainty into an agent's model of behavior for other agents.


The computation of the braid and collision probabilities is based on the forward simulation of system trajectory rollouts. For each path set $T\in\pazocal{T}_{i}$, agent $i$ considers the set of all control profiles drawn from $\mathbfcal{U}$, containing all combinations of control inputs corresponding to high and low magnitude speeds for all agents. For each pair $(T,U)\in T\times\mathbfcal{U}_{i}$, agent $i$ simulates a system trajectory $\Xi'$ by linearly projecting forward all agents from the current system state $\Xi(t)$ towards $T$ with a constant speed $U$. From each trajectory, it extracts a corresponding braid word $\beta_{i}$ (as described in Sec. \ref{sec:trajectoriesbraids}, using BraidLab \citep{braidlab}), and the minimum inter-agent distance $d_{min}$. By repeating this process for all rollouts, agent $i$ constructs a set $B\subset B_{n}$ comprising the set of possible braids that could be realizable in the remainder of the execution under the stated assumptions about agents' behaviors. Each braid $\beta^{*}\in B$ is then evaluated as:
\begin{equation}
P(\beta_{i} = \beta^*|\Xi,U,T) = 
\left\{
  \begin{array}{ll}
    1  & \mbox{if } \beta_{i}=\beta^* \\
    0 & \mbox{otherwise }
  \end{array}
\right.\mbox{.}
\end{equation}
This model acts as a switch that determines which rollouts should be considered for each braid found at the simulation stage. Finally, we model the probability of a collision $P(c|\Xi,U,T,\beta_{i})$ with the following sigmoid model:
\begin{equation}
P(c|\Xi,U,T,\beta_{i})=\frac{1}{1 + e^{a(d_{min}-\delta)}}\mbox{,}
\end{equation}
where $a$ controls the rate of change of the function and $\delta$ denotes a threshold distance beyond which collision is imminent. According to this model, the smaller $d_{min}$ is, it is exponentially more likely to have a collision.

\begin{figure*}
\centering
\begin{subfigure}{0.38\linewidth}
\centering
\includegraphics[trim = {2.3cm .5cm 3.3cm .9cm}, clip, width = \linewidth]{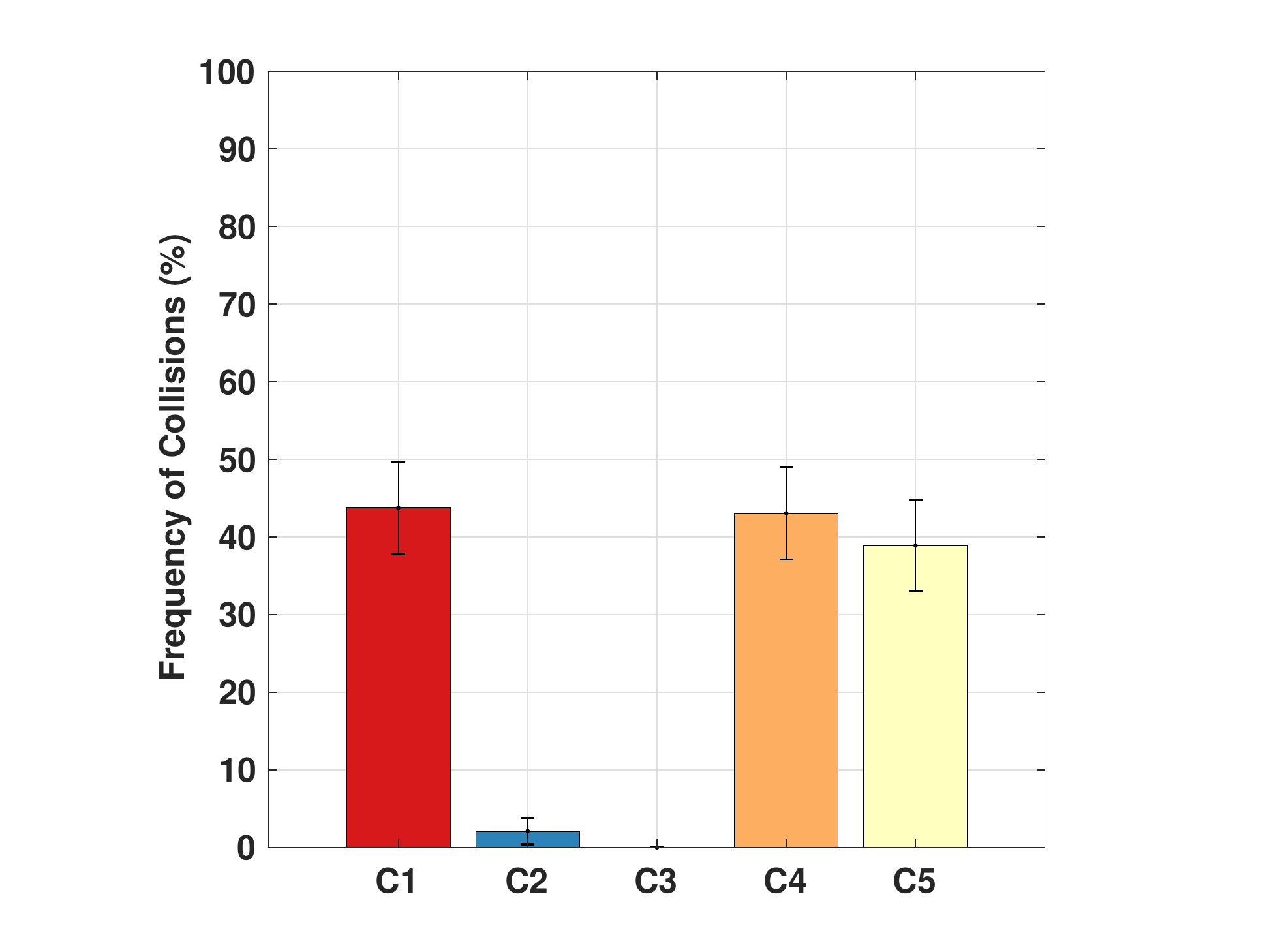}
\caption{Frequency of collisions for S1.\label{fig:numcol2agents}}
\end{subfigure}
~
\begin{subfigure}{0.38\linewidth}
\centering
\includegraphics[trim = {2.6cm .5cm 3.3cm .9cm}, clip, width = \linewidth]{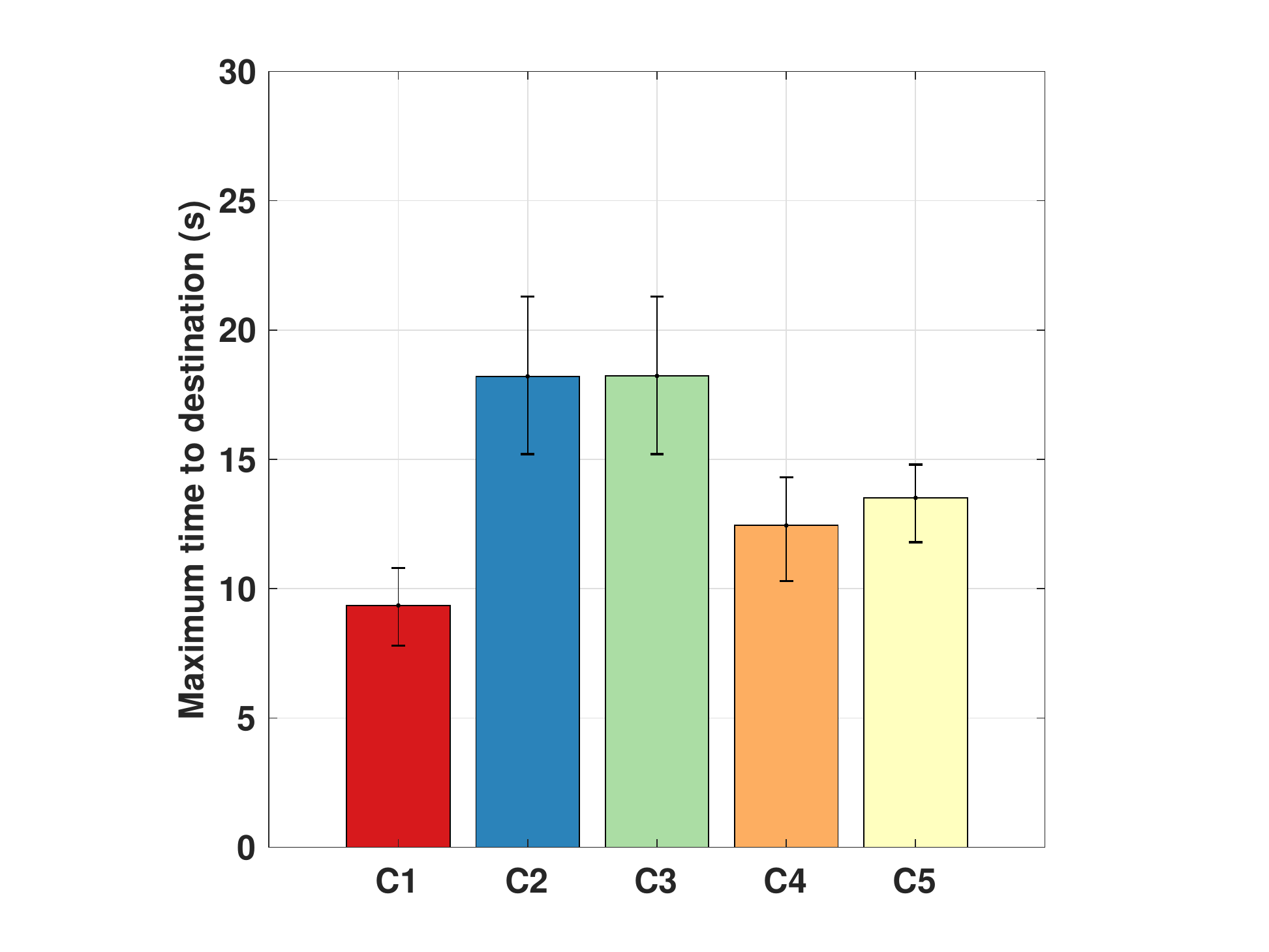}
\caption{Maximum time to destination for S1.\label{fig:time2agents}}
\end{subfigure}
\\
\begin{subfigure}{0.38\linewidth}
\centering
\includegraphics[trim = {2.3cm .5cm 3.3cm .9cm}, clip, width = \linewidth]{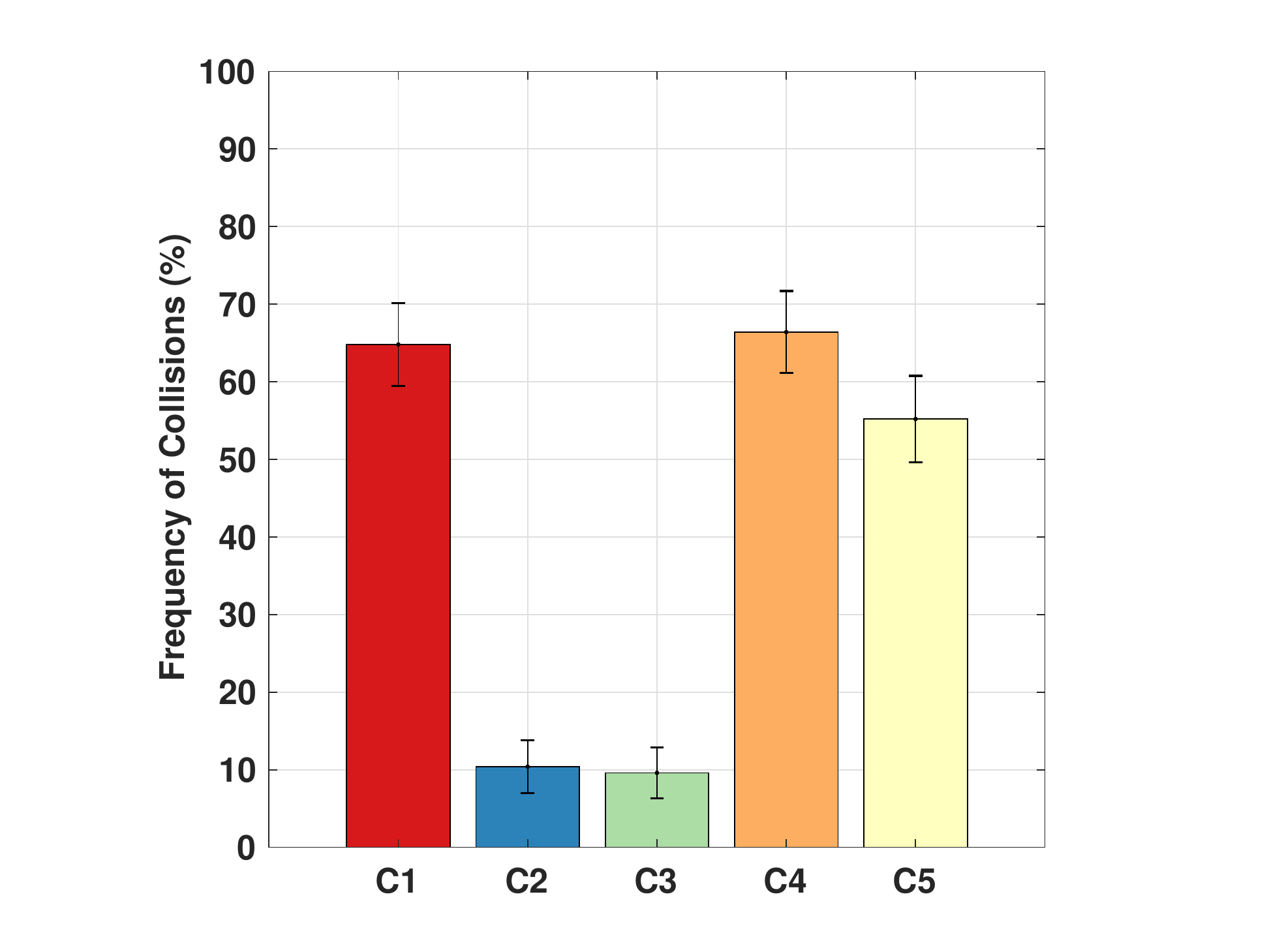}
\caption{Frequency of collisions for S2.\label{fig:numcol3agents}}
\end{subfigure}
~
\begin{subfigure}{0.38\linewidth}
\centering
\includegraphics[trim = {2.6cm .5cm 3.3cm .9cm}, clip, width = \linewidth]{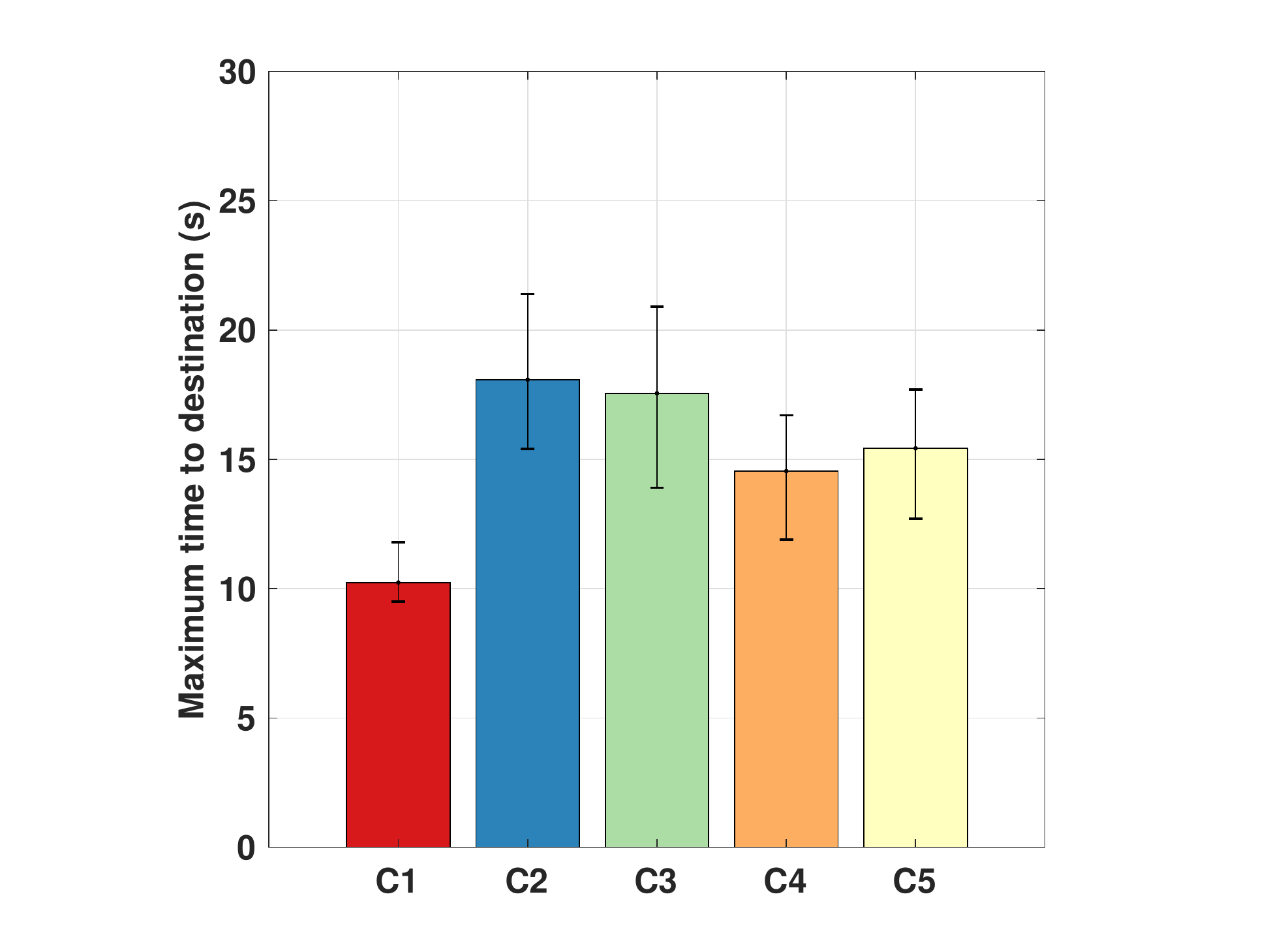}
\caption{Maximum time to destination for S2.\label{fig:time3agents}}
\end{subfigure}
\\
\begin{subfigure}{0.38\linewidth}
\centering
\includegraphics[trim = {2.3cm .5cm 3.3cm .9cm}, clip, width = \linewidth]{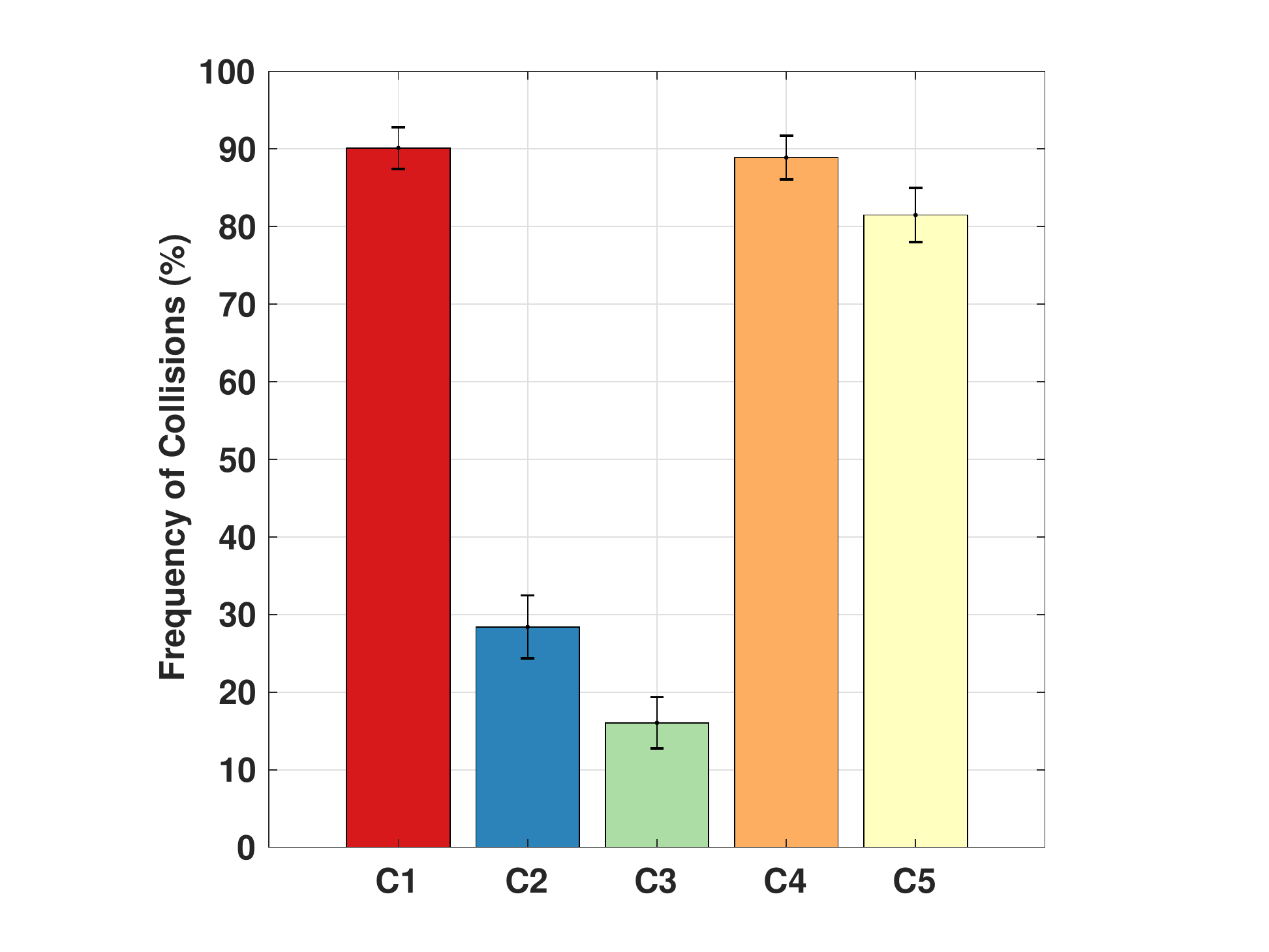}
\caption{Frequency of collisions for S3.\label{fig:numcol4agents}}
\end{subfigure}
~
\begin{subfigure}{0.38\linewidth}
\centering
\includegraphics[trim = {2.6cm .5cm 3.3cm .9cm}, clip, width = \linewidth]{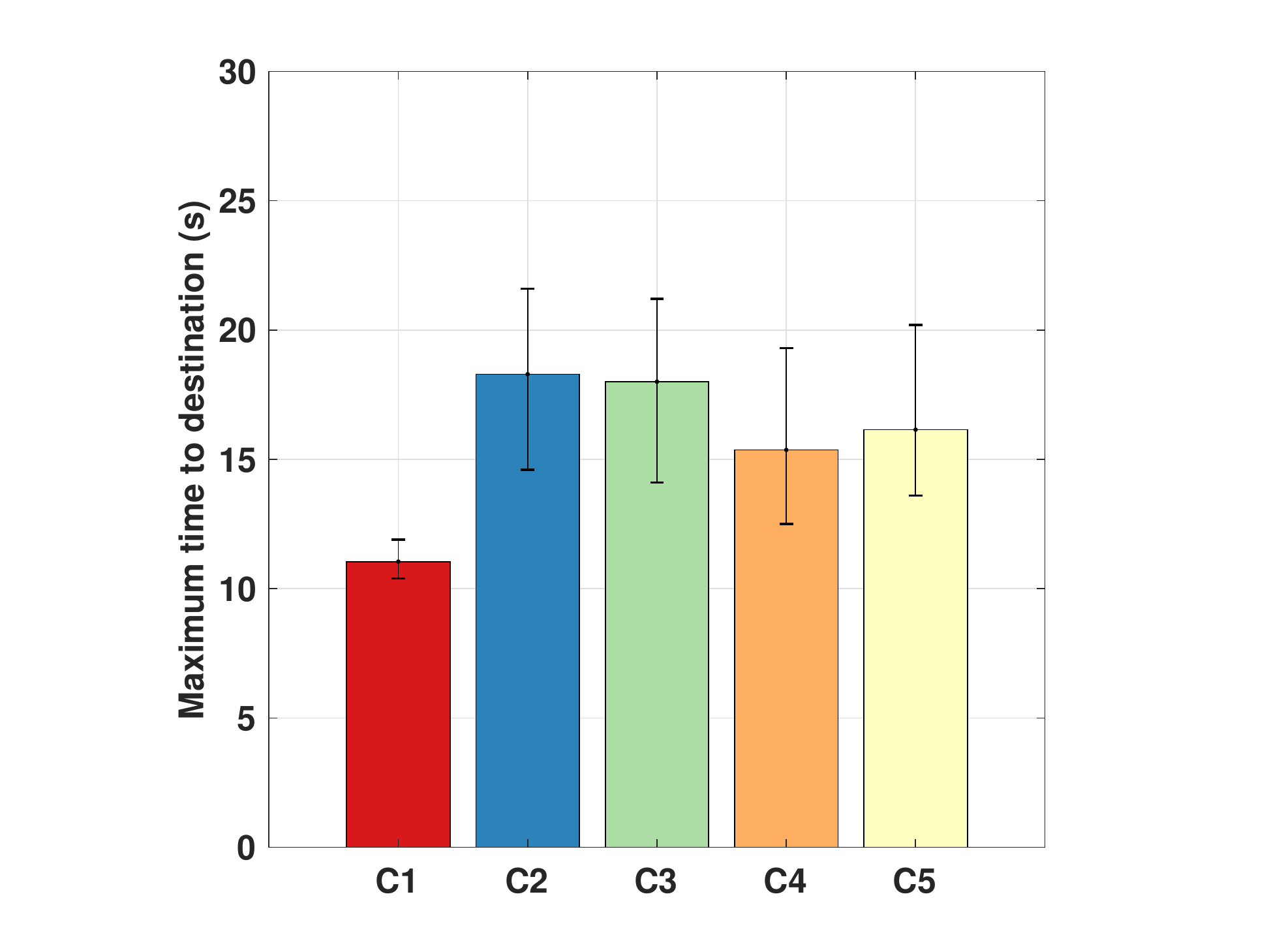}
\caption{Maximum time to destination for S3.\label{fig:time4agents}}
\end{subfigure}
\caption{Performance evaluation under homogeneous settings: (\subref{fig:numcol2agents}) and (\subref{fig:time2agents}) depict collision frequency and experiment time for S1 (2 agents), computed over 144 experiments; (\subref{fig:numcol3agents}) and (\subref{fig:time3agents}) depict collision frequency and experiment time for S2 (3 agents), computed over 125 experiments; (\subref{fig:numcol4agents}) and (\subref{fig:time4agents}) depict collision frequency and experiment time for S3 (4 agents), computed over 81 experiments. Bars correspond to conditions; error bars indicate standard deviations (assuming a collision is a Bernoulli event) and 25/75 percentiles in the collision frequency and time charts respectively.\label{fig:eval-homogeneous}}
\end{figure*}

\subsection{Evaluation I: Homogeneous Setting}\label{sec:eval-homogeneous}

We define three scenarios, involving 2, 3, and 4 agents respectively. For each scenario, we generate a set of experiments by varying agents' speed preferences. We execute each experiment under 5 conditions, each corresponding to a distinct variation of the proposed algorithm, executed by all agents (homogeneous setting). We then measure performance by looking at the frequency of collisions and the maximum experiment time per scenario and condition.

\textbf{Scenarios}:

\noindent S1: Two agents, starting from the bottom and the right sides of the intersection, are moving straight towards the top and left sides respectively. They both draw speeds from $\pazocal{U}_{s1}$ containing 12 evenly spaced speeds within $[5,10]$ ($m/s$). We generate 144 experiments corresponding to the Cartesian product $\pazocal{U}_{s1}^2$. 

\noindent S2: Three agents, starting from the bottom, right and top are moving straight towards the top, left and bottom sides respectively. They draw speeds from $\pazocal{U}_{s2}$, containing 5 evenly spaced speeds within the range $[5,10]$ ($m/s$). We generate 125 experiments corresponding to $\pazocal{U}_{s2}^3$. 

\noindent S3: Four agents, starting from the bottom, right, top, and left, are moving straight towards the top, left, bottom, and right sides respectively. They draw speeds from $\pazocal{U}_{s3}$, containing 3 evenly spaced speeds within the range $[5,10]$ ($m/s$). We generate 81 experiments corresponding to $\pazocal{U}_{s3}^4$. 

\begin{figure}
\centering
\begin{subfigure}{0.48\linewidth}
\centering
\includegraphics[trim = {3cm .1cm 3.2cm .1cm}, clip, width = \linewidth]{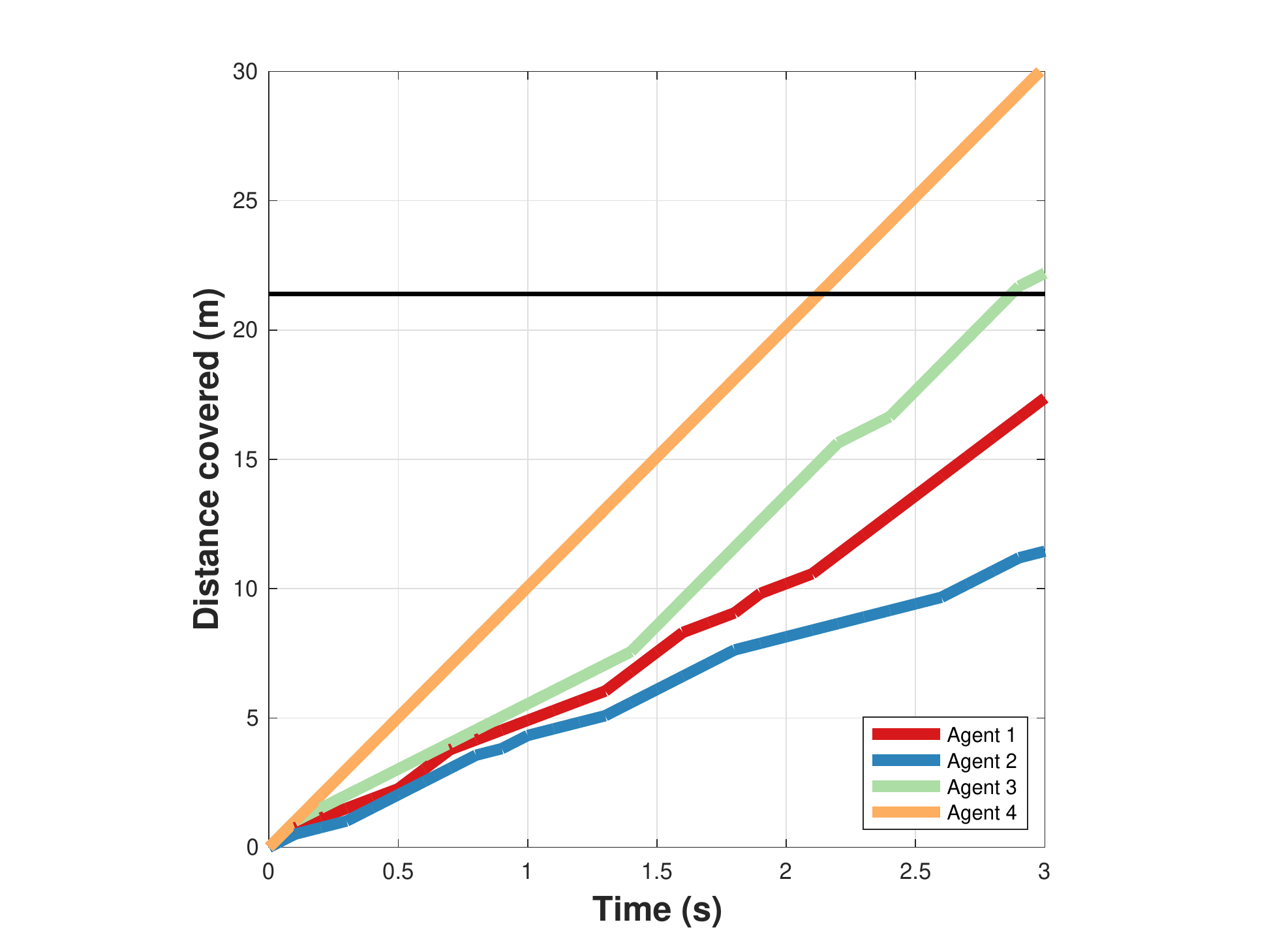}
\caption{Agents are running C2. \label{fig:C2}}
\end{subfigure}
~
\begin{subfigure}{0.48\linewidth}
\centering
\includegraphics[trim = {3cm .1cm 3.2cm .1cm}, clip, width = \linewidth]{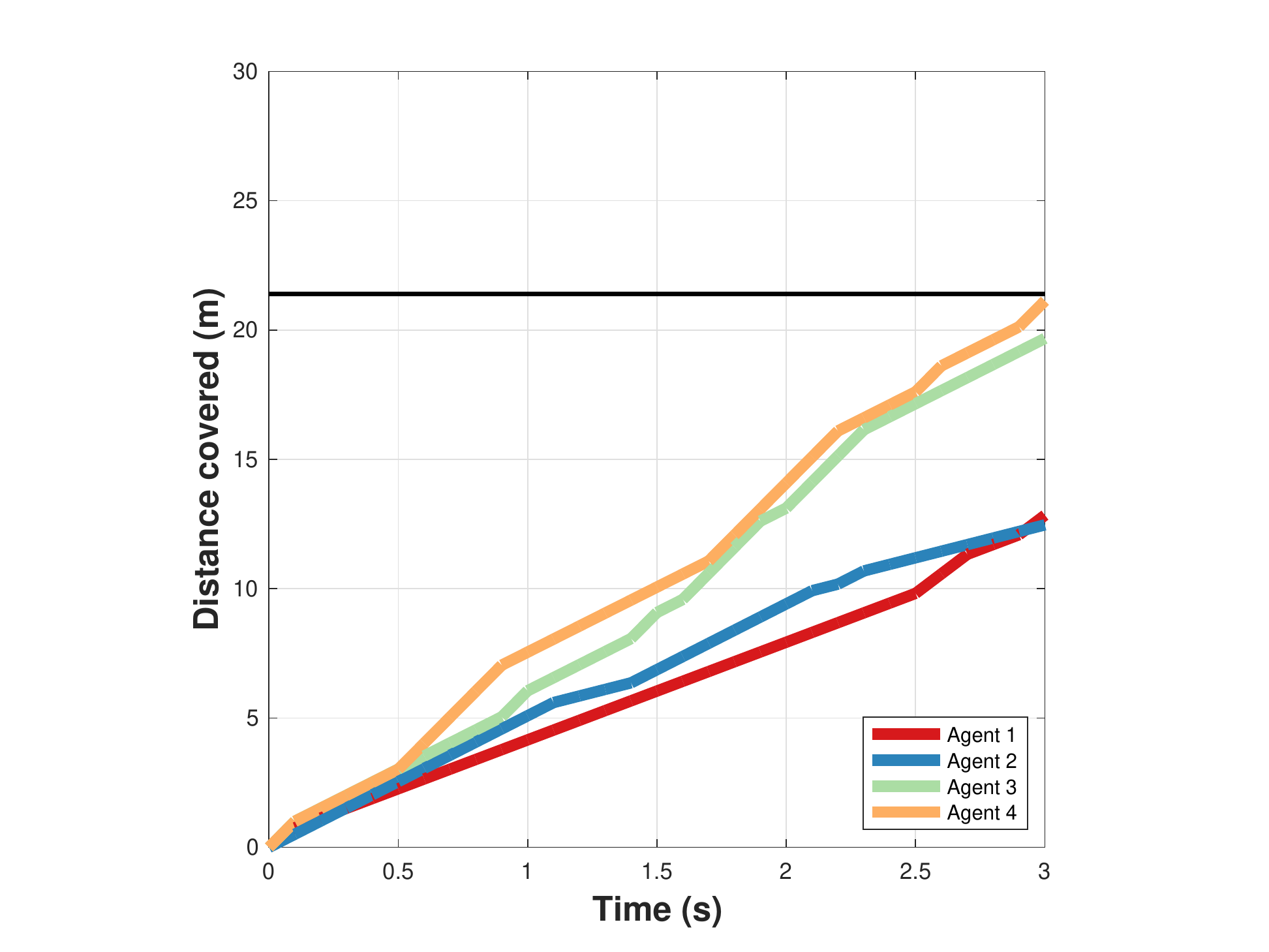}
\caption{Agents are running C4.\label{fig:C4}}
\end{subfigure}
\caption{Qualitative comparison: Distance covered per agent over the first $3s$ of execution within a 4-agent experiment. The black line indicates arrival at the intersection. We see that agents running C2 have reached a clear ordering of entry to the intersection, effectively corresponding to a consensus. In contrast, agents running C4 are clustered in pairs corresponding to an unclear ordering of intersection crossings.\label{fig:qualitative}}
\end{figure}

\textbf{Conditions}:

\noindent C1: An adaptation of the proposed scheme of eq.~\eqref{eq:spec} for which $w_{i} = 1$, i.e., agents track their desired paths with their desired speeds, without accounting for avoiding collisions with others. This condition serves as a characterization of the intensity of the multiagent encounters at the intersection for each scenario.

\noindent C2: The complete proposed algorithm.

\noindent C3: A modification of the proposed algorithm that assumes knowledge of the paths that other agents are following, i.e, they replace eq.~\eqref{eq:finalfinalbelief} with
\begin{equation}
bel(\tilde{b}_i) = \sum_{U}P(\tilde{{\beta}}_{i}|\Xi,U,T)P(U|\Xi,T)\mbox{.}
\end{equation}

\noindent C4: A variation of C2 that does not use braids for clustering trajectory sets. Specifically, agents reason about the emerging collision-free system trajectory $\tilde{\Xi}_{i}$ (instead of $\tilde{\beta}_{i}$), replacing eq. \eqref{eq:finalfinalbelief} with
\begin{equation}
bel(\tilde{\Xi}_{i}) = P(\tilde{\Xi}_{i}|\Xi,U,T)P(U|\Xi,T)P(T|\Xi)\mbox{.}\label{eq:c4}
\end{equation}

\noindent C5: A modification of C4 that assumes knowledge of the paths that others are following, i.e, C5 replaces eq. \eqref{eq:c4} with
\begin{equation}
bel(\tilde{\Xi}_{i}) = P(\tilde{\Xi}_{i}|\Xi,U,T)P(U|\Xi,T)\mbox{.}\label{eq:c5}
\end{equation}

\textbf{Analysis}:

\noindent Figure \figref{fig:eval-homogeneous} illustrates the performance of the selected algorithms across the three scenarios considered. As expected, C1 results in the highest collision frequency but lowest time to destination for all scenarios (red bars), serving as a characterization of the intensity of the selected scenarios. Our algorithm (C2) achieves consistently low collision frequencies for all scenarios (blue bars). Compared to C4, C2 reduces collision frequency by: 95\% across S1 (\figref{fig:numcol2agents}); 65\% across S2 (\figref{fig:numcol3agents}); 66\% across S3 (\figref{fig:numcol2agents}). C4, leveraging the knowledge of other agents' paths, consistently exhibits lower collision frequency than C5 across all scenarios. The price that C2, and C3 pay is the increased maximum time to destination; it can however be observed that for the more complex scenarios (S2, S3), the time difference is not significant (\figref{fig:time3agents}, \figref{fig:time4agents}). Note that a direct comparison of values across scenarios is not well defined as their parameter spaces (speed combinations) have different dimensionalities. We observe however that the general trends transfer across scenarios.



We interpret the performance gains as the result of effective incorporation of domain knowledge into decision making. The braid group represents the set of distinct modes that could describe the collective motion of navigating agents. Explicitly reasoning about these modes enables a rational agent to anticipate the effect of its actions on system behavior. Our policy outputs local actions of global outlook that contribute towards reducing uncertainty over the emerging mode. Collectively, this results in implicit coordination, reflected in the reduced collision frequency of C2, C3. To illustrate this point, \figref{fig:qualitative} depicts a comparative qualitative example of the behaviors generated by our policy. For the same experiment from S3 (run in the symmetric intersection of \figref{fig:notation}), we observe that C2 agents (\figref{fig:C2}) quickly converge to a clear order of intersection crossings as a result of their proactive decision making. On the other hand, C4 agents (\figref{fig:C4}), lacking the ability of modeling the complex multiagent dynamics, appear unable to coordinate their crossings and end up colliding. 

\subsection{Evaluation II: Reacting to Inattentive Agents}

Assume that agent 1 is an \emph{inattentive} agent that does \emph{not} account for collision avoidance in its decision making, i.e., navigates by solving eq.~\eqref{eq:spec} for $w_{1}=1$. In this section we investigate the effect of an inattentive agent's behavior to the performance of $n-1$ other rational agents, i.e., agents solving eq.~\ref{eq:spec} with $w_{i}\in (0,1)$, $i \neq 1$. To this end, we redefine the conditions C1-C5 from Sec.~\ref{sec:eval-homogeneous} as C1'-C5' by replacing agent 1 with an inattentive agent (C1' is identical to C1 --we include it as a characterization of the intensity of the scenarios considered). We execute the same scenarios S1-S3 from Sec.~\ref{sec:eval-homogeneous} under conditions C1'-C5'.

\begin{figure*}
\centering
\begin{subfigure}{0.38\linewidth}
\centering
\includegraphics[trim = {2.3cm .5cm 3.3cm .9cm}, clip, width = \linewidth]{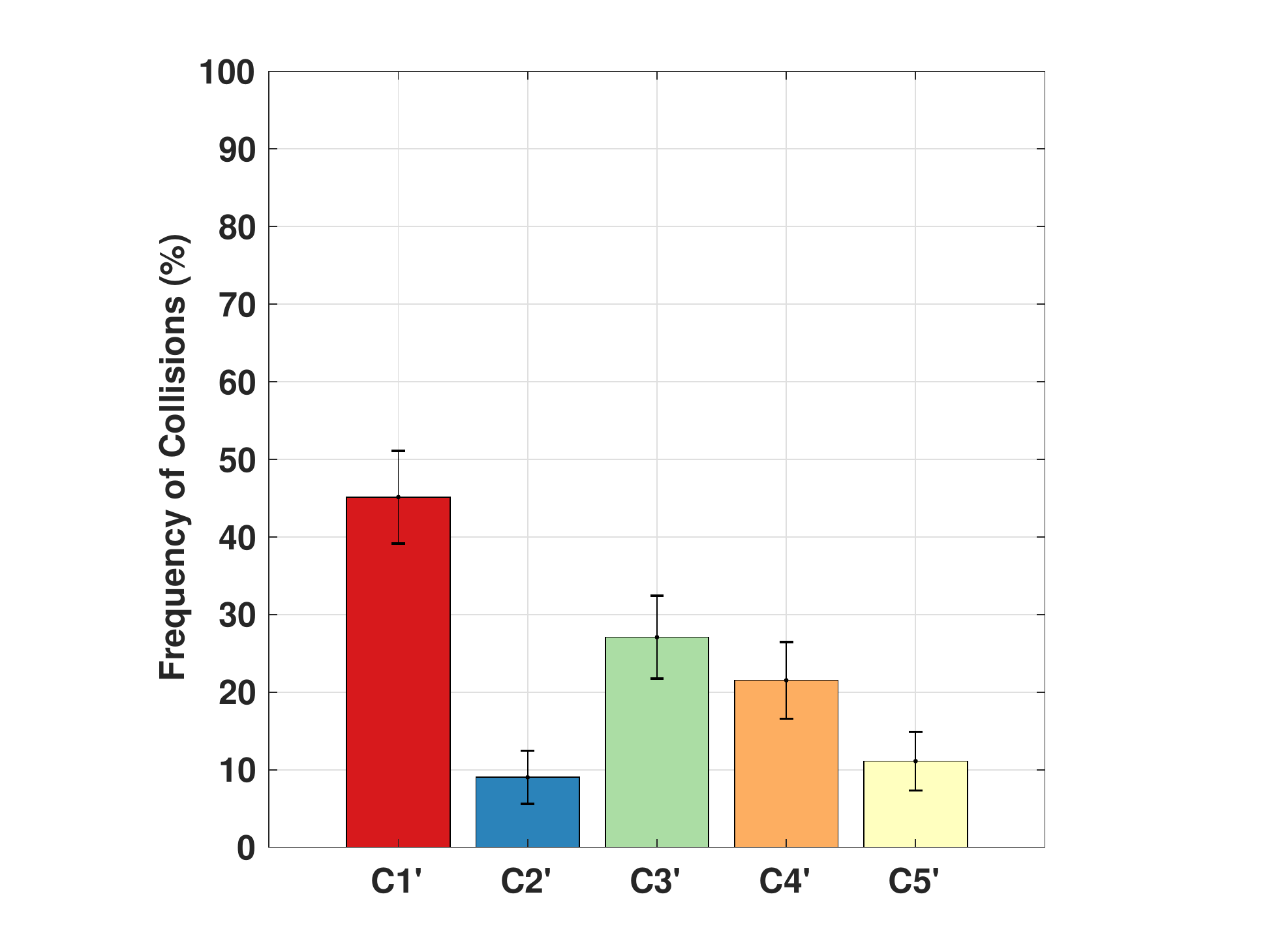}
\caption{Frequency of collisions for S1.\label{fig:numcol2agents-agent1antisocial}}
\end{subfigure}
~
\begin{subfigure}{0.38\linewidth}
\centering
\includegraphics[trim = {2.6cm .5cm 3.3cm .9cm}, clip, width = \linewidth]{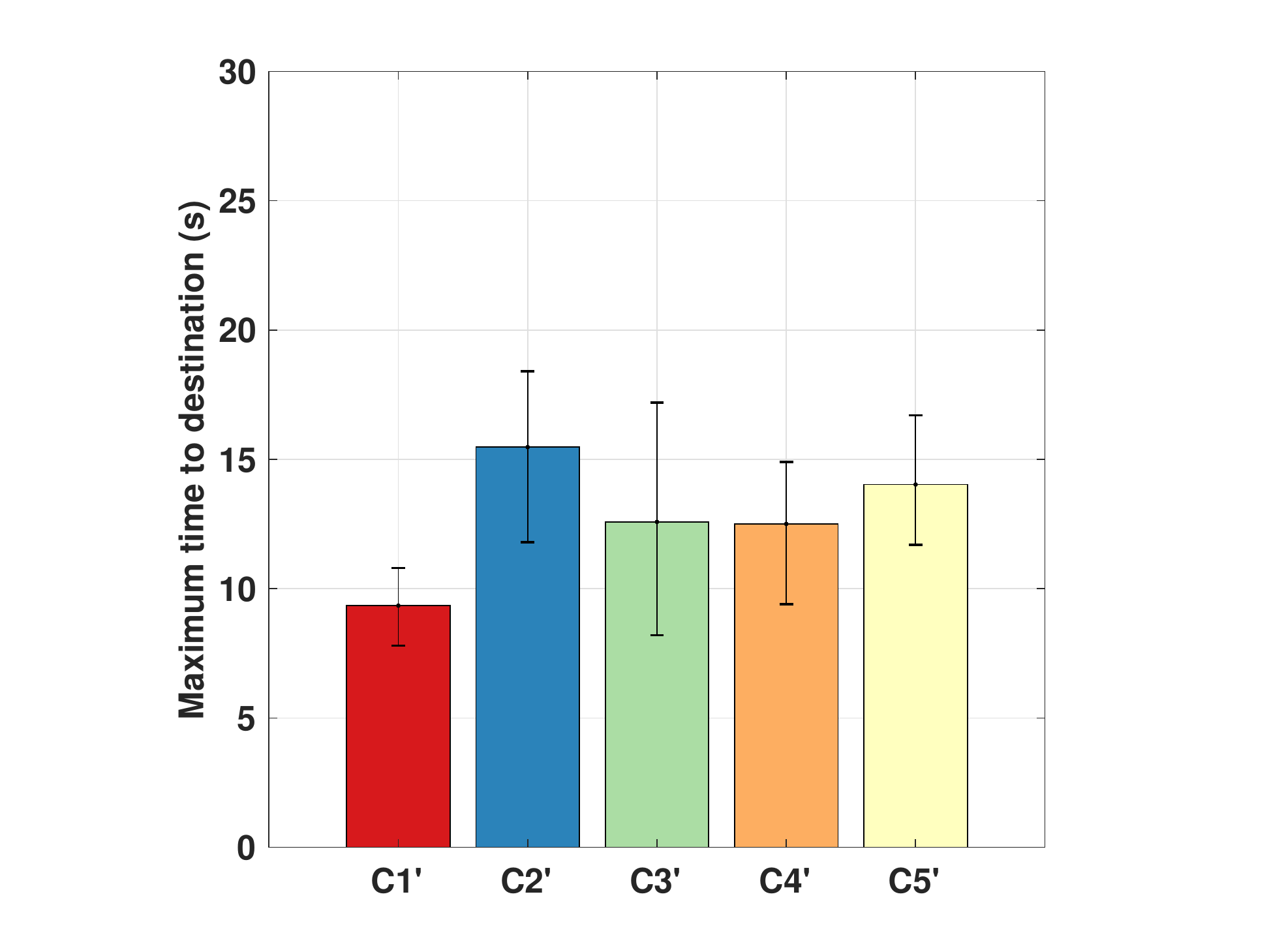}
\caption{Maximum time to destination for S1.\label{fig:time2agents-agent1antisocial}}
\end{subfigure}
\\
\begin{subfigure}{0.38\linewidth}
\centering
\includegraphics[trim = {2.3cm .5cm 3.3cm .9cm}, clip, width = \linewidth]{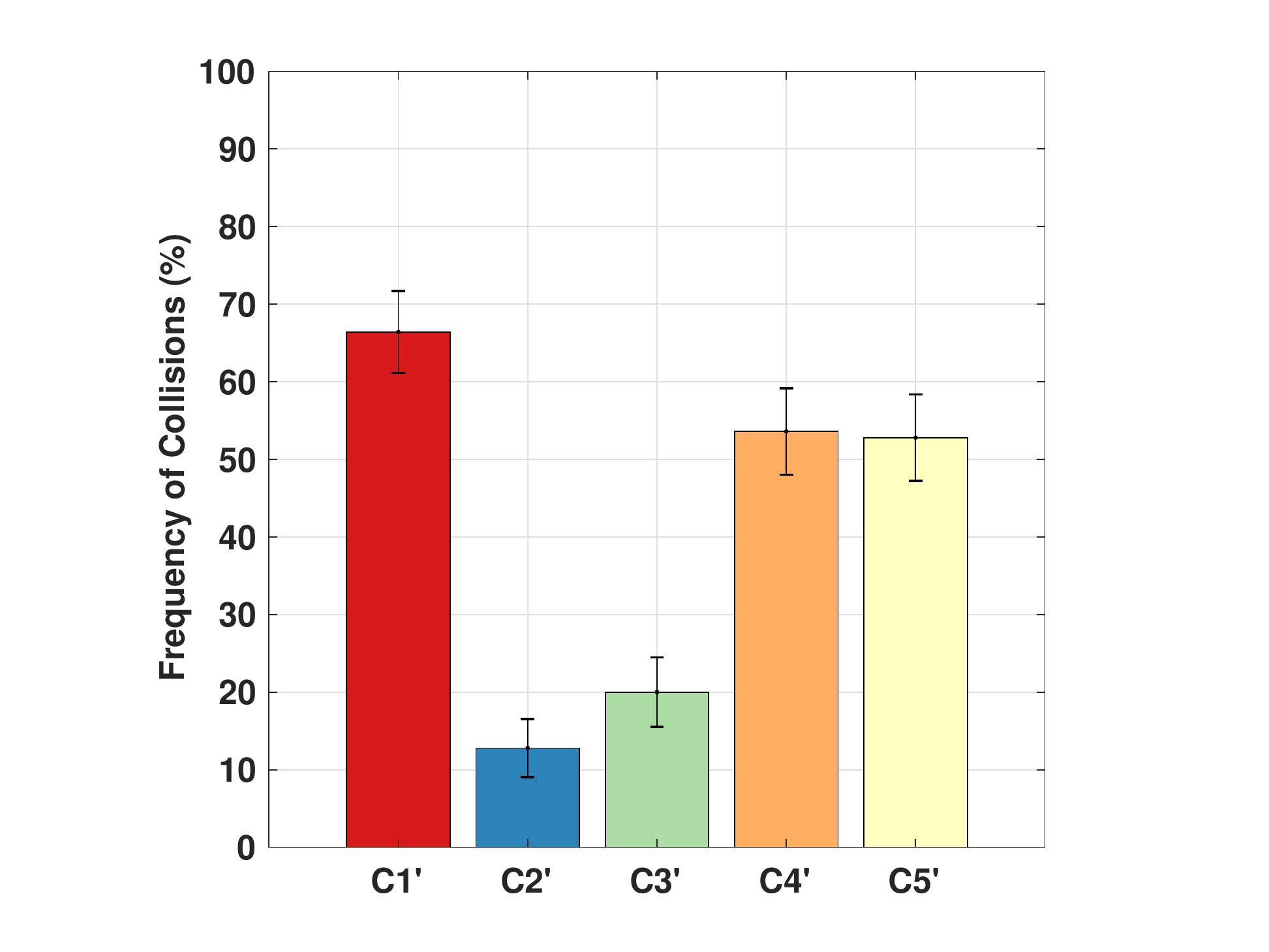}
\caption{Frequency of collisions for S2.\label{fig:numcol3agents-agent1antisocial}}
\end{subfigure}
~
\begin{subfigure}{0.38\linewidth}
\centering
\includegraphics[trim = {2.6cm .5cm 3.3cm .9cm}, clip, width = \linewidth]{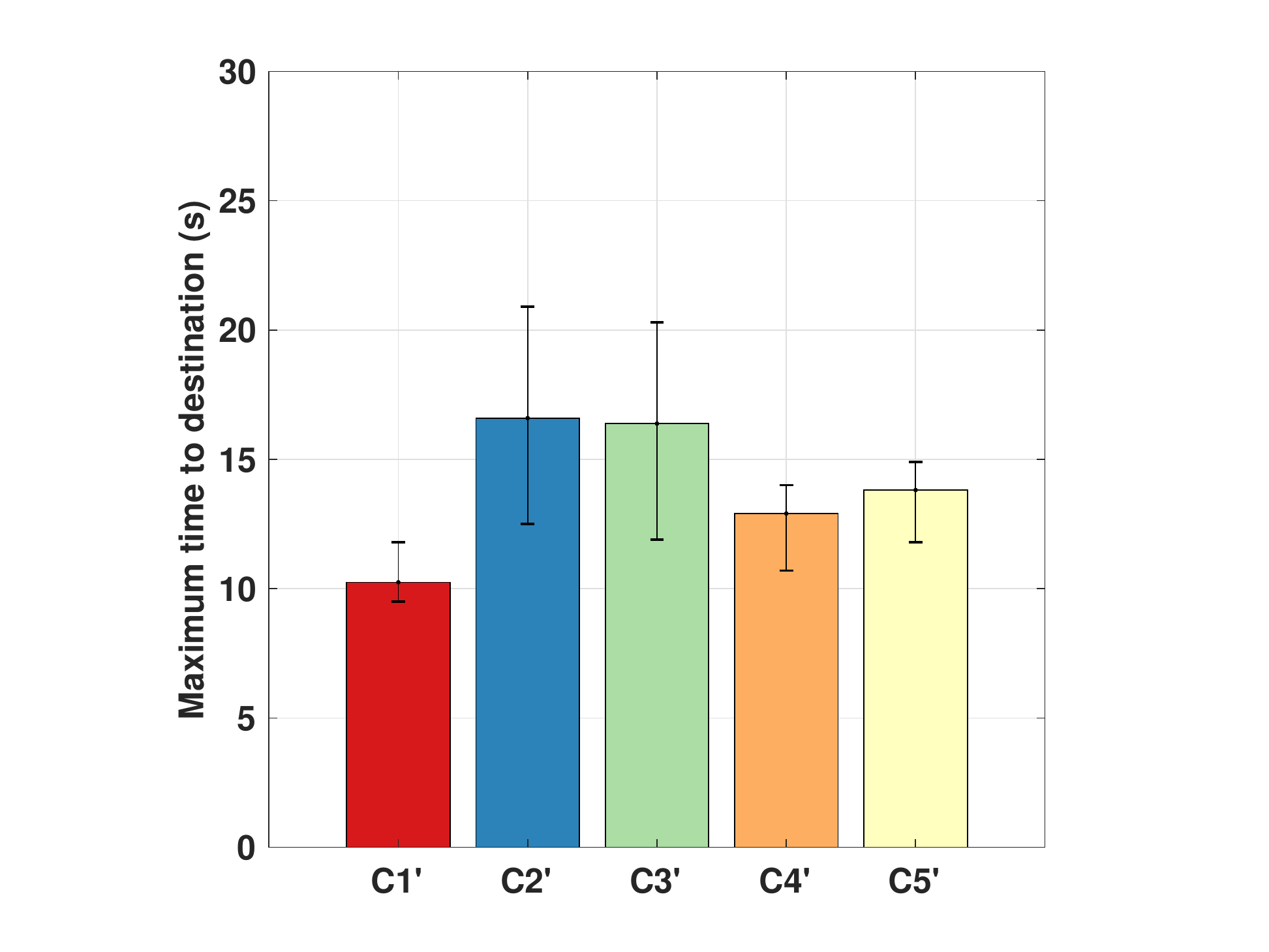}
\caption{Maximum time to destination for S2.\label{fig:time3agents-agent1antisocial}}
\end{subfigure}
\\
\begin{subfigure}{0.38\linewidth}
\centering
\includegraphics[trim = {2.3cm .5cm 3.3cm .9cm}, clip, width = \linewidth]{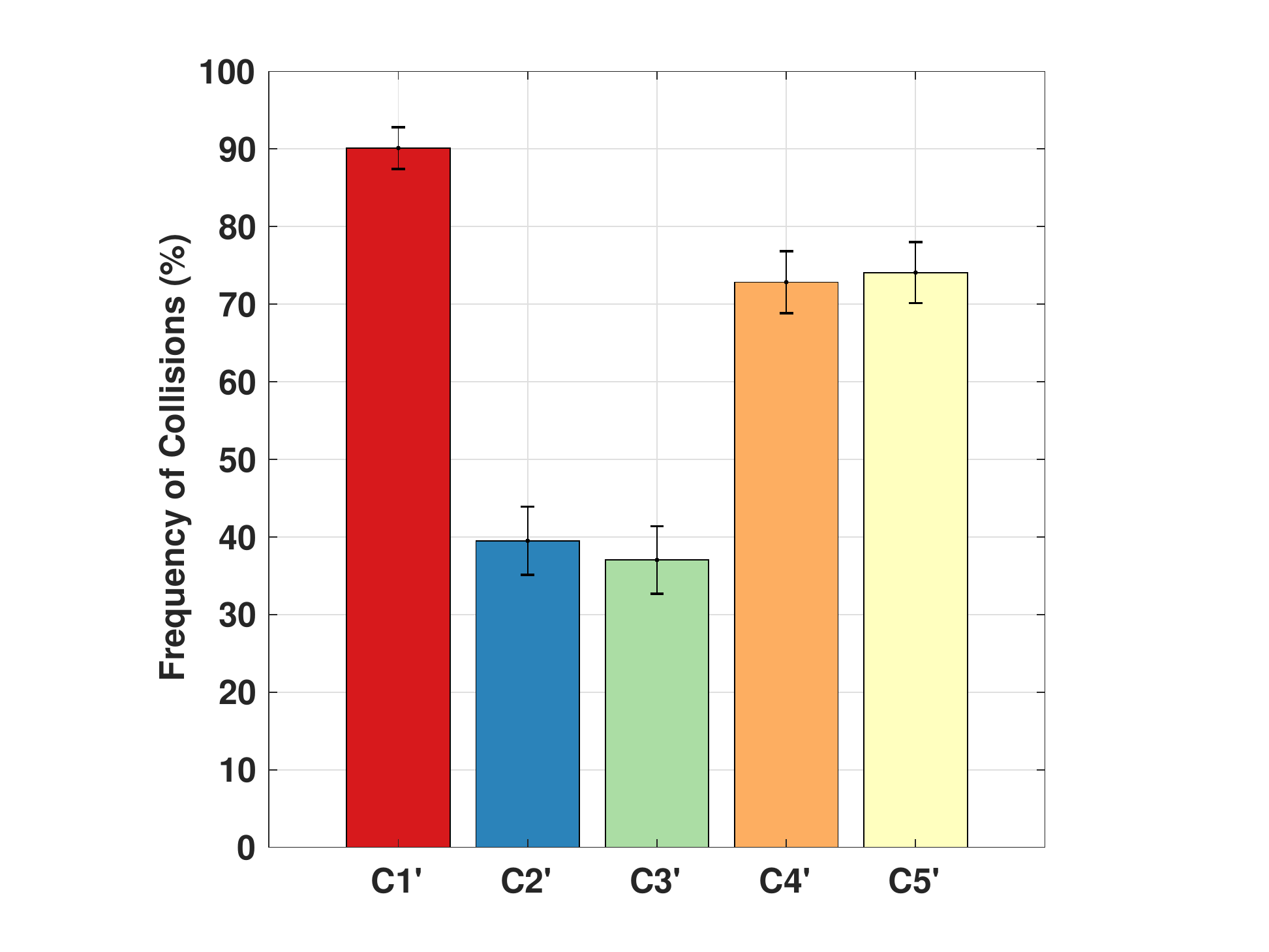}
\caption{Frequency of collisions for S3.\label{fig:numcol4agents-agent1antisocial}}
\end{subfigure}
~
\begin{subfigure}{0.38\linewidth}
\centering
\includegraphics[trim = {2.6cm .5cm 3.3cm .9cm}, clip, width = \linewidth]{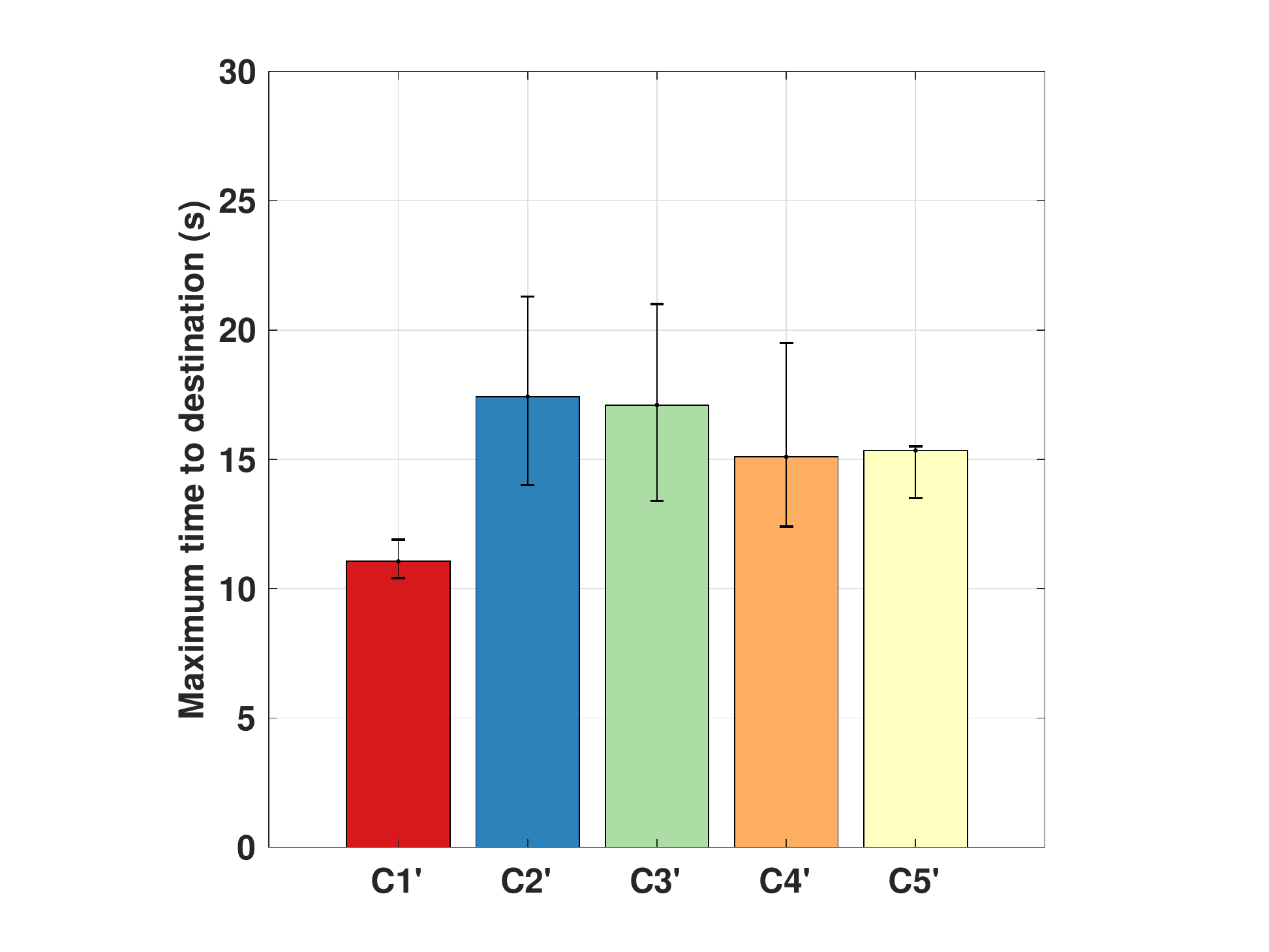}
\caption{Maximum time to destination for S3.\label{fig:time4agents-agent1antisocial}}
\end{subfigure}
\caption{Performance evaluation under heterogeneous settings: (\subref{fig:numcol2agents-agent1antisocial}) and (\subref{fig:time2agents-agent1antisocial}) depict collision frequency and experiment time for S1 (2 agents), computed over 144 experiments; (\subref{fig:numcol3agents-agent1antisocial}) and (\subref{fig:time3agents}) depict collision frequency and experiment time for S2 (3 agents), computed over 125 experiments; (\subref{fig:numcol4agents-agent1antisocial}) and (\subref{fig:time4agents-agent1antisocial}) depict collision frequency and experiment time for S3 (4 agents), computed over 81 experiments. Bars correspond to conditions; error bars indicate standard deviations (assuming collision is a Bernoulli event) and 25/75 percentiles in the collision frequency and time charts respectively.\label{fig:eval-heterogeneous}}
\end{figure*}

\textbf{Analysis}:

\figref{fig:eval-heterogeneous} depicts our findings from running the described scenarios under the new conditions. Overall, we see the majority of the trends from Sec.~\ref{sec:eval-homogeneous} to transfer in this heterogeneous domain. Non-topological methods (C4', C5') tend to perform better in terms of time efficiency than the topological ones (C2', C3'), whereas the latter perform better in terms of collision frequency. The distinction in collision frequency is exemplified as the number of agents (and thus the scenario complexity) increases. While we see that the overall scenario complexity increase results in overall decline in performance across all conditions, C2' and C3' are responding to it better, consistently achieving less than half collisions than C4' and C5' in S2 and S3. This behavior illustrates the ability of our framework to robustly handle less predictable scenarios involving agents that violate the rationality assumption as set by eq.~\eqref{eq:spec}.

The patterns of \figref{fig:eval-heterogeneous} also exhibit a few interesting differences compared to the patterns of \figref{fig:eval-homogeneous}, stemming from the introduction of the inattentive agent. In particular, we see that time efficiency is improved overall --the fact that the inattentive agent leaves the intersection with its preferred high speed results in an emptier workspace faster. We also see that overall C2' and C3' perform comparably, but note a trend of C3' scoring higher than C2' in terms of collision frequency for S1 and S2, despite the fact that it is aware of other agents' destinations. We believe that this might have to do with the fact that C2, considering more topological outcomes is able to react more cautiously to the unpredictable behavior of the inattentive agent.

\section{Discussion}\label{sec:discussion}

We presented a planning framework for decentralized navigation in structured multiagent domains with no communication, such as unsignalized intersections. In such environments, the geometric structure of the environment often constrains the trajectories of rational agents to belong to a finite set of modes. Observing that these modes bear distinct topological properties, we employed the formalism of topological braids \citep{birman} to abstract them into symbols. This abstraction enabled us to construct a probabilistic inference mechanism that predicts the emerging topological mode given past agents' behaviors. Based on this mechanism, we designed a cost-based planner that generates actions that are uncertainty-reducing over the space of future topological modes. By running our planner, agents execute actions that collectively prune unsafe and unlikely outcomes, accelerating uncertainty reduction and effectively yielding convergence to a collision-avoidance protocol. This is reflected in executions of significantly lower collision frequency compared to a series of baselines reasoning directly over the space of trajectories, across a series of challenging experiments at a simulated unsignalized street intersection domain. We showed that this finding transfers to scenarios involving inattentive agents, demonstrating the robustness of our framework in handling uncertainty over other agents' behavioral models.

Our findings may have broader implications about the value of topological features for multiagent navigation. Reasoning over a bounded set of modes could effectively enable significant complexity reduction compared to reasoning directly over the space of multiagent Cartesian trajectories. For reference, from the perspective of the ego-agent, the space of possible 4-agent trajectories over an horizon of $H = 10$ time steps, assuming a control space $\pazocal{U}^4 = 10^4$ has size $S_{t} = |\pazocal{T}_{i}||\pazocal{U}^4|^{H} = 27 \cdot 10,000^{10}$. The space of braids that could be practically possible for any $n$-agent scenario could be upper-bounded\footnote{Note that this computation may include duplicate braids due to simplifications happening between consecutive generators (e.g., $\sigma_{1}\cdot\sigma_{1}^{-1} = \sigma_{0}$).} to $S_{b} \leq [2(n-1)]^{D}+1$ under the assumption of a maximum depth $D$ representing the maximum number of generators appearing in a braid word describing the execution. For a 4-agent scenario with $D = 5$ (the depth appearing the most across our 4-agent experiments), this number would be $S_{b} = 7,777$. Note that in practice, the size of the space of relevant topological outcomes may even be significantly smaller than $S_{b}$ depending on the problem setup and agents' state history.

\subsection{Limitations}

Although braids have the potential of significantly compressing the space of outcomes, and thus relaxing inference, in this paper we did not leverage the projected computation gains, as we conditioned our belief on the control profile and the system path (see eq. \eqref{eq:finalbelief}). Ongoing work involves learning a distribution over the space of braids from a dataset of intersection scenarios. Reasoning \emph{directly} over braids during execution will enable the outlined computation speedups and allow for scaling to more complex scenes.

Furthermore, although the considered scenario captures the main features of an unsignalized intersection, the setup is deliberately simplified to facilitate the extraction of foundational insights. Moving forward, we plan on incorporating more high-fidelity behavioral models into the decision-making processes of other agents and considering additional scenarios involving pedestrians. We also plan on validating our approach with real-world hardware experiments on a miniature robotic racecar \citep{srinivasa2019mushr} at an indoors street intersection mockup.

Finally, our evaluation setup was based on an ablation study, specifically chosen to illustrate the benefits of incorporating topological features in the inference mechanism. Although we did not compare our approach against baselines from the literature, we see our framework as a significant \emph{complement} and extension of alternative approaches. Topological features could augment and improve the performance of existing belief-space approaches \citep{Bouton17}, reinforcement learning techniques \citep{Isele17} or prosocial control frameworks \citep{Sadigh2018, Lazar_CDC18}.

\bibliographystyle{plainnat}
\bibliography{references}

\balance

\end{document}